\RequirePackage{fix-cm}
\documentclass[natbib,twocolumn]{svjour3}          
\smartqed
\journalname{IJCV}
\usepackage{graphicx}
\usepackage{algorithm}
\usepackage{algorithmicx}
\usepackage[bottom]{footmisc}
\usepackage[export]{adjustbox}
\usepackage[noend]{algpseudocode}
\usepackage[svgnames]{xcolor}
\usepackage[utf8]{inputenc}
\usepackage{acronym}
\usepackage{adjustbox}
\usepackage{afterpage}
\usepackage{amsfonts}
\usepackage{amsmath}
\usepackage{amsopn}
\usepackage{amssymb}
\usepackage{array}
\usepackage{array}
\usepackage{array}
\usepackage{bbm}
\usepackage{booktabs}
\usepackage{colortbl}
\usepackage{color}
\usepackage{epigraph}
\usepackage{float}
\usepackage{graphics}
\usepackage{graphics}
\usepackage{lscape}
\usepackage{makeidx}
\usepackage{mathtools}
\usepackage{mathtools}
\usepackage{multirow}
\usepackage{natbib}
\usepackage{nicefrac}
\usepackage{pifont}
\usepackage{subcaption}
\usepackage{tabularx}
\usepackage{type1cm}         %
\usepackage{url}
\usepackage{xparse}
\usepackage{xspace}
\usepackage{xstring}
\usepackage[defaultlines=0]{nowidow}
\definecolor{tealblue}{rgb}{0.18, 0.40, 0.46}
\definecolor{ruby}{rgb}{0.88, 0.07, 0.37}
\usepackage[pagebackref=true,breaklinks=true,citecolor=tealblue,colorlinks,linkcolor=ruby,bookmarks=false]{hyperref}
\algdef{SE}[DOUNTIL]{Do}{doUntil}{\algorithmicdo}[1]{\algorithmicuntil\ #1}

\newcommand{\gain}[1]{\textcolor[rgb]{0.18, 0.40, 0.86}{#1}}

\begin{document}

	\title{Dynamic Curriculum Learning\\ for Great Ape Detection in the Wild}
	\titlerunning{Dynamic Curriculum Learning for Great Ape Detection in the Wild}

	\author{
		Xinyu Yang \and
		Tilo Burghardt \and
		Majid Mirmehdi
	}

	\institute{
		Xinyu Yang $^1$ \\ \email{xinyu.yang@bristol.ac.uk}      \vspace{3pt} \\    
            Tilo Burghardt $^1$\\ \email{tilo@cs.bris.ac.uk}     \vspace{3pt} \\    
            Majid Mirmehdi $^1$\\  \email{majid@cs.bris.ac.uk} \vspace{5pt} \\    
				$^1$ Department of Computer Science, University of Bristol, UK
	}
	\sloppy
\maketitle
\begin{abstract}
    We propose a novel end-to-end curriculum learning approach for sparsely labelled animal datasets leveraging large volumes of unlabelled data to improve supervised species detectors. We exemplify the method in detail on the task of finding great apes in camera trap footage taken in challenging real-world jungle environments. In contrast to previous semi-supervised methods, our approach  adjusts learning parameters dynamically over time and gradually improves detection quality by steering training towards virtuous self-reinforcement. To achieve this, we propose integrating pseudo-labelling with curriculum learning policies and show how learning collapse can be avoided. We discuss theoretical arguments, ablations, and significant performance improvements against various state-of-the-art systems when evaluating on the Extended PanAfrican Dataset holding approx.~$1.8M$ frames. We also demonstrate our method can outperform supervised baselines with significant margins on sparse label versions of other animal datasets such as Bees and Snapshot Serengeti. We note that performance advantages are strongest for smaller labelled ratios common in ecological applications. Finally, we show that our approach achieves competitive benchmarks for generic object detection in MS-COCO and PASCAL-VOC indicating wider applicability of the dynamic learning concepts introduced. We publish all relevant source code, network weights, and data access details for full reproducibility.
\end{abstract}

\section{Introduction}

\begin{figure*}[ht]
	\centering
	\includegraphics[width=0.8\textwidth]{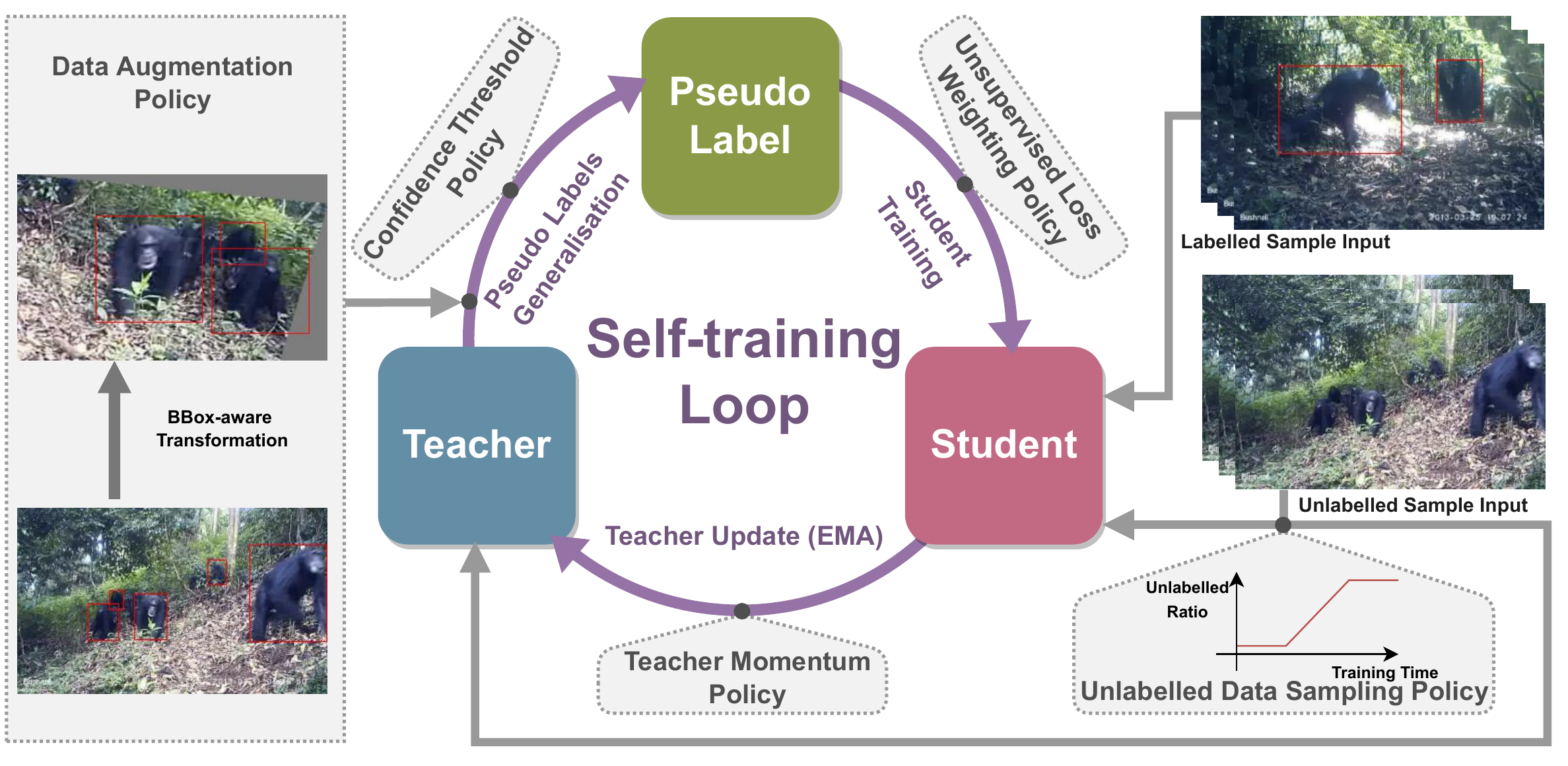}
	\caption{\footnotesize{
			\textbf{Conceptual Overview.} {We utilise a student-teacher paradigm for learning where the teacher produces pseudo-labels for the student to learn from while being updated  by an exponentially moving average (EMA) of the student model. We apply five dynamic policies to this learning loop that we show can lead to effective (i.e. virtuous) self-training cycles:  \emph{unlabelled data sampling policy} to control the unlabelled sample input,  \emph{confidence threshold policy} to filter unreliable pseudo-labels, \emph{data augmentation policy} to diversify the unlabelled training data, \emph{unsupervised loss weighting policy} to balance unsupervised and supervised losses, and \emph{teacher momentum policy} to adjust the update speed of the teacher model.
	}}}
	\label{fig:concept}
\end{figure*}

{\textbf{Motivation --} Automated visual monitoring of animals filmed in their natural habitats is gaining significant traction, boosted recently by a plethora of deep learning methods and applications {\citep{tabak2019machine,norouzzadeh2021deep,tuia2022perspectives}}.
However, developing and advancing relevant computer vision tools remains challenging due to several factors. Animals in their natural environments are often hard to detect, obscured by dynamic backgrounds, varying illumination conditions, occlusions, camouflage effects, and more. Deploying network models trained on prevalent image and video databases, such as ImageNet \citep{deng2009imagenet}, MS-COCO \citep{lin2014microsoft}, Kinetics \citep{carreira2017quo}, are often insufficient on their own, even after taking advantage of the potentials of transfer learning. To further exacerbate the difficulty of deploying machine learning methods to their fullest extent in the domain, there is still a distinct lack of large-scale, annotated training datasets for particular species despite evolving general frameworks~\citep{megadetector}. Whilst crowd sourcing annotations can help, low labelling rates relative to archive sizes remain the norm in the field. For great apes in particular, several recent works have attempted to address some of the above mentioned challenges~\citep{yang2019great,schofield2019chimpanzee,sakib2020visual,doi:10.1126/sciadv.abi4883}. However, these works still either only pretrain on datasets from other domains or rely on relatively small datasets for supervised training due to the complexities associated with obtaining annotations. Thus, while these methods have advanced the cause somewhat regarding great ape detection in jungle settings, they have also emphasised the urge for better use of the vast archives of completely unlabelled camera trap footage.}

\noindent{\textbf{Paper Concept --} In response, this paper introduces a novel curriculum learning approach that intertwines traditional supervised detector training with unlabelled data utilisation. The approach demonstrates by proof-of-concept that, exemplified for great apes, large unlabelled camera trap archives can indeed be exploited to enrich and empower real-world animal detector construction without any further labelling efforts. We leverage lessons learned from recent self-supervised~\citep{grill2020bootstrap, caron2021emerging,chen2021exploring} and semi-supervised~\citep{sohn2020simple,sohn2020fixmatch,xu2021end} methods on feature representation learning and image classification to propose an end-to-end student-teacher based detection pipeline that integrates self-training~(via pseudo-labels) and dynamic training polices into one cyclical curriculum learning design. Our model learns from unlabelled data in the curriculum by generating high quality pseudo-labels on the fly. In turn, these virtual annotations of otherwise unlabelled samples are exploited by the student whose update influences the teacher and a next round of pseudo-label generation. This cyclical self-training idea can be illustrated conceptually as a learning loop shown in Fig.~\ref{fig:concept}. We will demonstrate that carefully fine-tuned curriculum learning policies in this loop can blend labelled and unlabelled sample input in a way that leads to virtuous training cycles (as opposed to vicious training cycles) which increasingly and consistently improve model performance. Critically, we show that dynamic learning adjustments can be controlled stably by policies and can improve performance over static learning. Intuitively, the approach expands model coverage of the vast space of animal appearance in particular, slowly from the labelled sample base, guided and channelled by the policies. We show that this approach can significantly improve great ape detection benchmarks, {as well as other benchmarks including Bees and Snapshot Serengeti}. We also demonstrate that the method is applicable beyond the targeted {animal} domain and achieves competitive {or state-of-the-art} results on the MS-COCO {and PASCAL-VOC} object detection challenges without a need for dataset-specific hyperparameter fine-tuning.

\noindent{\textbf{Contributions --} Overall, the contributions of this paper can be summarised as, {(i)} a novel end-to-end \emph{dynamic} detection framework for semi-supervised curriculum learning designed to improve species detectors built from sparsely labelled datasets,
{(ii)} a dynamic policy system with stable hyper-parameters for temporal control over changing learning properties in semi-supervised detector training promoting self-reinforcing virtuous training loops, (iii) extensive experiments and ablations on a large scale real-world great ape camera trap dataset - we report improvements to the state-of-the-art  for the semi-supervised great ape detection task evaluated on the Extended PanAfrican Dataset, {(iv) we offer new semi-supervised detection benchmarks on sparse labelling versions of two other animal datasets - Bees and Snapshot Serengeti, contributing towards handling annotation shortage in the animal domain,} and finally (v) we also provide competitive and state-of-the-art semi-supervised object detection results for the MS-COCO {and PASCAL-VOC} datasets, demonstrating broader applicability.}}

\section{Related Work}
In this section, we consider works related to the key topics of interest with focus on the state-of-the-art.

\noindent\textbf{Semi-supervised Learning (SSL) -- } SSL exploits the potential of unlabelled data to facilitate model learning with limited amounts of annotated data {\citep{Rebuffi20SSL}}. Training computer vision models such as objection detection or action recognition networks, relies on the availability of annotated datasets which can be costly to generate. This has motivated the development of semi-supervised methods {\citep{jeong2019consistency,berthelot2019mixmatch,zhai2019s4l,sohn2020fixmatch,sohn2020simple,zhang2021flexmatch,xu2021end,tang2021humble}}.

One dominant SSL approach is consistency regularisation where the model is regularised to generate consistent predictions on data with different augmentations \citep{jeong2019consistency,berthelot2019mixmatch,zhai2019s4l}. Another approach is based on generating pseudo-labels for unlabelled data and updating the model by training on a mix of unlabelled data with pseudo-labels and labelled data with manually-annotated labels  \citep{sohn2020fixmatch,sohn2020simple,zhang2021flexmatch,xu2021end,tang2021humble,liu2021unbiased}.   What type of pseudo-labelling to use is critical to the success of SSL in particular scenarios. FixMatch \citep{sohn2020fixmatch} applied a high confidence threshold for mining pseudo-labels and then these sharpened and strongly-augmented pseudo-labels were utilised for model training. STAC \citep{sohn2020simple} extended FixMatch from image classification to objection detection by introducing self-training and augmentation-driven consistency regularisation. More recently, \cite{xu2021end} introduced the soft teacher mechanism to alleviate the issue of unreliable pseudo-labels generated by the teacher in SSL object detection. {\cite{liu2021unbiased} jointly train a student and a teacher in a mutually-beneficial manner by applying a class-balance loss to down-weight overly confident pseudo-label impact.}
{In the light of the success of these methods, our approach follows the pseudo-labelling concept, but addresses the model learning challenges differently.}

\begin{figure*}[th]
	\centering
	\includegraphics[width=0.99\textwidth]{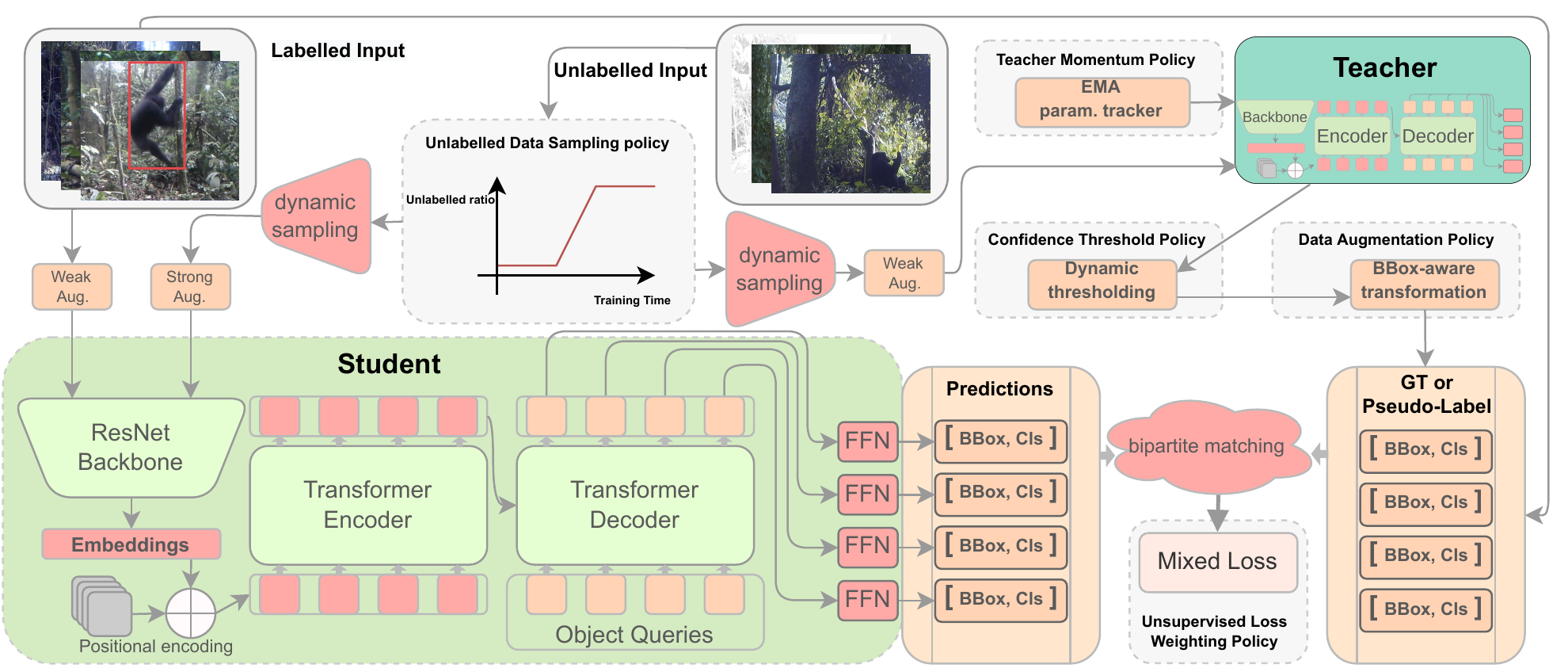}
	\caption{\footnotesize{
			{\textbf{Detailed End-to-end Self-training Great Ape Detection Pipeline.} We utilise the Deformable DETR \citep{zhu2020deformable} framework with a ResNet backbone as detector architecture. The student network (light green) uses this architecture as well as the teacher network (dark green). All labelled data along with dynamically sampled and policy-controlled unlabelled data are mixed during training. The teacher performs pseudo-label generation with purely unlabelled input on the fly. The pseudo-labels are filtered with an adaptive threshold and then augmented via a bounding box-aware transformation. The teacher network is updated by student model via a dynamic momentum coefficient. The final loss is the sum of supervised and unsupervised detection losses balanced by a policy-controlled dynamic weight. We carefully designed the policies of the system to achieve an effective (a.k.a. virtuous) self-reinforcing training cycle.}
	}}
	\label{fig:overview}
\end{figure*}
{\noindent\textbf{Object detection - } This area of computer vision has advanced in leaps and bounds since the very start of the modern era of deep learning. Some notable early works are: (i) single-stage detection frameworks, such as \citep{redmon2016you,liu2016ssd,lin2017focal,tian2019fcos}, which perform object classification and bounding box regression directly, without using pre-generated region proposals. They are typically applied over a dense sampling of possible object locations to estimate the class probabilities and bounding box coordinates directly. (ii) In contrast, two-stage detection frameworks, such as  \citep{ren2015faster,he2017mask,lin2017feature} utilise a region proposal network to generate class-agnostic regions of interest (ROIs) and only then perform ROI bounding box regression and object classification.}
More recently, DEtection with TRansformers (DETR) \citep{carion2020end} built the first end-to-end detection pipeline by viewing object detection as a direct set-prediction problem. DETR eliminated the need for anchor-based target assignment pre-processing and non-maximum suppression (NMS) post-processing, prevalent in commonly used object detectors. It combined CNNs for feature extraction and transformers for feature interpretation to directly translate object queries to class and bounding boxes by leveraging cross attention \citep{vaswani2017attention} on image features. However, the vanilla DETR suffers from slow convergence and hence longer training time than detectors based on YOLO, SSD and Faster-RCNN. The Deformable DETR \citep{zhu2020deformable} proposed a deformable attention module that only attend to a small set of prominent key elements to replace the attention in DETR. This improvement led to faster convergence and a better performance. We select this variant as our model for the various detection components of our proposed curriculum learning framework.

\noindent\textbf{Curriculum Learning (CL) -- }
The CL training approach \citep{bengio2009curriculum,surveycurriculum} has had significant impact on the design of computer vision algorithms, such as  \citep{karras2017progressive,wang2018weakly,huang2020curricularface,wang2022pseudo,zhang2021flexmatch}. \cite{wang2018weakly}, for instance, use average precision of each sample to re-rank the data from easy to hard and train the object detector in an easy-to-hard fashion applied to the pre-ranked order of data. \cite{wang2022pseudo} propose a pseudo-labelled auto-curriculum learning framework that engages reinforcement learning to learn a series of dynamic thresholds for the pseudo-labels for semi-supervised key-point localisation.
FixMatch \citep{sohn2020fixmatch}, on the other hand, applied a constant threshold to select unlabelled samples for training, which fails to address the learning difficulties at different time steps. Thus, it can allow poor quality samples to get through.
FlexMatch \citep{zhang2021flexmatch} improved on FixMatch  by dynamically  adjusting the threshold at each time step to filter unlabelled samples and pseudo-labels.

Both FixMatch and FlexMatch applied a pre-trained pseudo-label generator which does not get updated during the semi-supervised learning stage, thus, failing to consider the evolution of the pseudo-label generator as the learning progressing. To address this issue, we propose student-teacher learning paradigms inspired by the recent advances in self-supervised learning methods \citep{grill2020bootstrap,chen2021exploring,caron2021emerging} that evolve the teacher component dynamically guided by a set of curriculum learning policies and controls. We will now describe our approach in detail.

\section{Proposed Method}

We introduce an end-to-end curriculum learning pipeline for effective semi-supervised Great Ape detection in camera trap footage. Our framework follows a student-teacher training scheme, as illustrated in detail in Fig.~\ref{fig:overview} and operates as follows: in each learning iteration, we train a student model built around a Deformable DETR detector \citep{zhu2020deformable}   by a mix of labelled and unlabelled videos where the unlabelled videos are sampled by our curriculum sampling policy $\pi$.
The teacher performs pseudo-label generation with unlabelled input.  Pseudo-labels are then refined by a dynamic threshold $\varsigma_t$ and transformed by augmentation policy $\mathcal{A}$. Together, both the pseudo-labels and manually-annotated labels are fed into the student network for learning.
The student network is then updated by the gradient from the overall loss which is balanced by unsupervised loss weight $\alpha_t$. Finally, the teacher network is updated by the exponential moving average (EMA) of the student parameters via a dynamic momentum coefficient $m_t$.
This completes one iteration of the learning loop leading to an updated teacher and student model. Our target will be to design the mentioned policies and an appropriate loss in a fashion that virtuous, that is effective, learning can be practically achieved.

\subsection{Problem Definition}
Let us consider that frames are sampled from the video at a frequency of $\omega$ for both labelled and unlabelled videos. The teacher is trained to generate the pseudo-labels for unlabelled frames only, while the student is trained to fit the pseudo-labels with the unlabelled input frames, as well as the ground-truth labels with the labelled input frames. Thus, the overall loss for the student is defined as the weighted sum of supervised and unsupervised losses:
\begin{equation}\label{equ:lossall}
	\footnotesize{
		\mathcal{L}_{all}=\mathcal{L}+\alpha\mathcal{L}^{\prime},
	}
\end{equation}
where $\mathcal{L}$ and $\mathcal{L}^{\prime}$ denote the supervised loss of labelled samples and unsupervised loss of unlabelled samples respectively, and $\alpha$ represents the balancing weight.

\begin{figure*}[ht]
	\centering
	\includegraphics[width=0.85\textwidth]{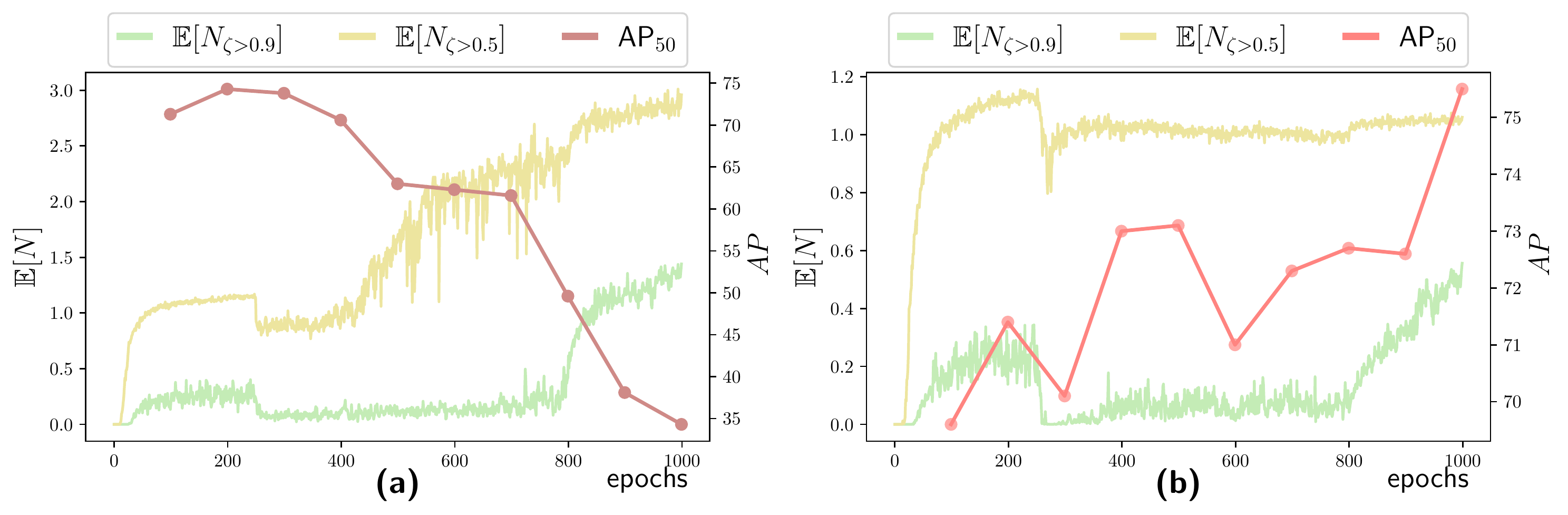}
	\caption{\footnotesize{{\textbf{Bipolar Behavioural Dynamics of Learning via Self-training Loops}. Two representative cases illustrating the bipolar dynamics of training success when using the proposed architecture: \textit{vicious} collapse in \textbf{(a)} and \textit{virtuous} effective learning in \textbf{(b)}. The scenarios differ only in policy  parameterisation{\protect\footnotemark}. In both plots, the right ordinate indicates the $AP_{50}$ on the validation set whilst the left ordinate represents the average number of samples with confident score $\zeta > 0.9$ or $\zeta >0.5$.}}}
	\label{fig:bipolarisation}
\end{figure*}

Further consider we have a labelled sample set $D_X$ with $M$ labelled samples $X_i$ and their corresponding class and box labels $(C_i, B_i)$ and an unlabelled sample set $D_{U}$ with $N$ unlabelled samples $U_j$ (regardless of the sampling approach) used for training. Also, let $\mathfrak{S}(\theta)$ be the student model parameterised by $\theta$, and $\mathfrak{T}(\theta^{\prime})$ be the teacher model parameterised by $\theta^{\prime}$. Eq. (\ref{equ:lossall}) can then be expanded to:
\begin{equation}\label{equ:lossfunc}
	\footnotesize{
		\begin{split}
			\mathcal{L}_{all}(\theta)= & \mathcal{L}+\alpha\mathcal{L}^{\prime} \\
			= & \frac{1}{M} \sum_{X_i \in D_X} L_{\theta}\left(X_{i}\right)+\alpha\frac{1}{N}\sum_{U_j \in D_{U}} L_{\theta}^{\prime}\left(U_j\right) ~,
		\end{split}
	}
\end{equation}
where  $L_{\theta}\left(X_{i}\right)$ is the loss for labelled sample $X_i$,
\begin{equation}
	\footnotesize{
		\begin{split}
			L_{\theta}\left(X_{i}\right)=&\operatorname{Loss} \big( \mathfrak{S}\left(X_{i},{\theta}\right), \left[C_{i},B_i \right] \big) \\
			=&L_{reg}\big(\mathfrak{S}\left(X_{i},{\theta}\right),B_i\big)+L_{ce}\big(\mathfrak{S}\left(X_{i},{\theta}\right),C_i\big) ~,
		\end{split}
	}
\end{equation}
\noindent and $L_{\theta}^{\prime}\left(U_j\right)$ is the loss for the unlabelled sample $U_j$,
\begin{equation}
	\footnotesize{
		\begin{split}
			L_{\theta}^{\prime}\left(U_j\right)=&\operatorname{Loss}
			\big(\mathfrak{S}\left(U_j,{\theta}\right), \mathfrak{T}\left(U_j,{\theta^{\prime}}\right) \big) ~, \\
			=&L_{reg}\big(\mathfrak{S}\left(U_j,\theta\right),\mathfrak{T}\left(U_j,\theta^{\prime}\right)\big) \\
			 &+L_{ce}\big(\mathfrak{S}\left(U_j,\theta\right),\mathfrak{T}\left(U_j,\theta^{\prime}\right)\big)  ~,
		\end{split}
	}
\end{equation}
where $L_{reg}$ represents the bounding box regression loss and $L_{ce}$ represents the classification loss.

We follow common practice in self-supervised learning methods, such as  \citep{caron2021emerging,grill2020bootstrap}, so that the teacher is updated by the EMA of the student,
\begin{equation} \label{equ:momentum}
	\footnotesize{
		\theta^{\prime}_{t} \leftarrow m\theta^{\prime}_{t-1}+(1-m)\theta_{t} ~.
	}
\end{equation}
Our objective is to find a set of student parameters $\theta^*$ that minimises the expected overall loss $\mathcal{L}_{all}(\theta)$, such that
\begin{equation}
	\footnotesize{
		\begin{array}{c}
			\theta^*=\underset{\theta}{\arg \min } \mathcal{L}_{all}(\theta) ~.
		\end{array}
	}
\end{equation}

\subsection{Self-reinforcing Training Loop}

\begin{figure*}[ht]
	\centering
	\includegraphics[width=0.75\linewidth]{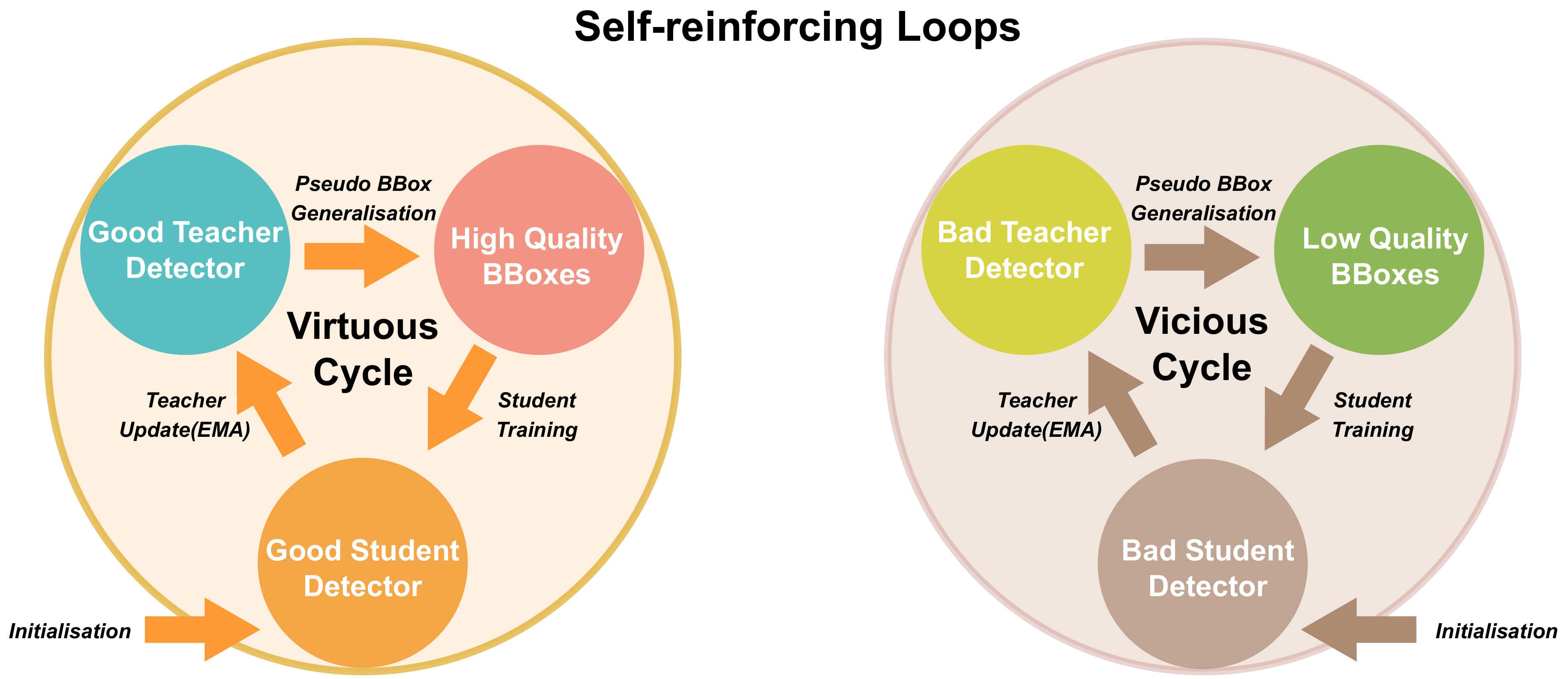}
	\caption{\footnotesize{
			{\textbf{Processes within Self-reinforcing Loops}. Illustration of four key processes (arrows) involved in training loops. Note that a destabilised \textit{Virtuous Cycle} where low quality pseudo labels or highly inaccurate student or teacher networks are produced turns into a \textit{Vicious Cycle} and vice versa. Thus, effective parameterisations and policies for the key processes are required to to promote stable learning. 
	}}}
	\label{fig:circle}
\end{figure*}

The evolution of the student and teacher network is conceptually a cyclic relationship. On the one hand, the performance of the student detector depends on the quality of the pseudo-labels, which in turn relies on the teacher, and on the other hand, the teacher is updated according to student status. Thus, there is an intricate interdependence between the student, the teacher, and the pseudo-labels forming a self-training loop which is controlled by the learning policies.

\footnotetext{For reproducibility of behaviour in Fig. \ref{fig:bipolarisation}, the exact parameters used were: (a) $\pi$: constant policy, constant $\varsigma=0.1$ and constant $\alpha=0.1$; (b) $\pi$: linear increase policy,  linear increase $\varsigma ~0.3\to 0.5$, constant $\alpha=0.5$. Both were trained for 1000 epochs with the first 250 epochs for warmup, $lr$ decreases at 800$^{th}$ epoch.}

In practice, we observe a bipolarisation phenomenon for the training of models with different settings where they  gradually become more confident of their predictions, but show two drastically different performance trajectories. As illustrated on sample training runs shown in Fig. \ref{fig:bipolarisation}, whilst a gradual increase of confidence indicators $\mathbb{E}[N_{\zeta>0.9}]$ and $\mathbb{E}[N_{\zeta>0.5}]$, which represent  the average number of predicted objects whose confident scores $\zeta$ are over 0.9 or 0.5  respectively, can be observed; the validation performance  can be erratic, either collapsing or improving effectively.
In the example illustrated in Fig. \ref{fig:bipolarisation}(a),  a decrease of $AP_{50}$ on the validation set was observed after a few hundred training epochs. Initially, one may assume the model is simply over-fitting at this stage in learning. However, as shown in the second example in Fig. \ref{fig:bipolarisation}(b) for a different parameterisation, a long-term increase in $AP_{50}$ can be observed, which suggests the model can learn from the training set well into training cycles. We hypothesise that the bipolar collapse or success of learning with regard to generalisation is critically linked to the self-reinforcement property of the training loop parameterisation and policies.

We categorise bipolarisation as two different types of learning cycles, effective \emph{Virtuous Cycles} and collapsing  \emph{Vicious Cycles} as illustrated in
Fig. \ref{fig:circle}. In the virtuous cycle state, the teacher model generates pseudo-labels of sufficient quality as to contribute to the training of the student model, allowing both models to  improve continually. In contrast, the vicious cycle sees the teacher generate insufficiently low quality pseudo-labels that degrade the training of the student model, thus both models degenerate continually.

Fig. \ref{fig:circle} depicts four key processes in the learning loop (shown as arrows), which are crucially influencing the trajectory of learning:
(i) \emph{Initialisation}: initialising the student model before the self-training phase;
(ii) \emph{Teacher Update}: updating the teacher network according to student status;
(iii) \emph{Pseudo-label Generalisation}: generating pseudo-labels by teacher;
(iv) \emph{Student Training}: using pseudo-labels to update student. Our goal is to find suitable controls that guide the above processes and can maintain the development of a virtuous self-training loop and, for robustness, also transition from a vicious to a virtuous setting.

For initialisation, we confirmed experimentally that the proposed system operates in a stable manner with fixed, standard backbone initialisations across all tested datasets. In particular, we use the self-supervised ImageNet pre-trained ResNet weights from SWAV~\citep{caron2020unsupervised} for our detection backbone. In addition, to show that general training stability can also be maintained in a supervised initialisation scenario, we test supervised ImageNet pre-trained ResNet~\citep{he2016deep} weights too. Note that any such fixed initialisation is essential as random initialisation triggers vicious training cycles, however, the fix is not sensitive to target dataset properties as transfer between scenarios still produces stable learning~(see Section 7).

For the other three processes above, we propose appropriate `policies' that guide learning within the confounds of effective 'virtuous' learning cycles - these are described next.


\subsection{Student Training} \label{studentintervention}
{Student network training is guided by two policies that allow the student model to exploit the unlabelled sample data and their pseudo-labels effectively.}

\begin{figure*}[ht]
	\centering
	\includegraphics[width=0.75\linewidth]{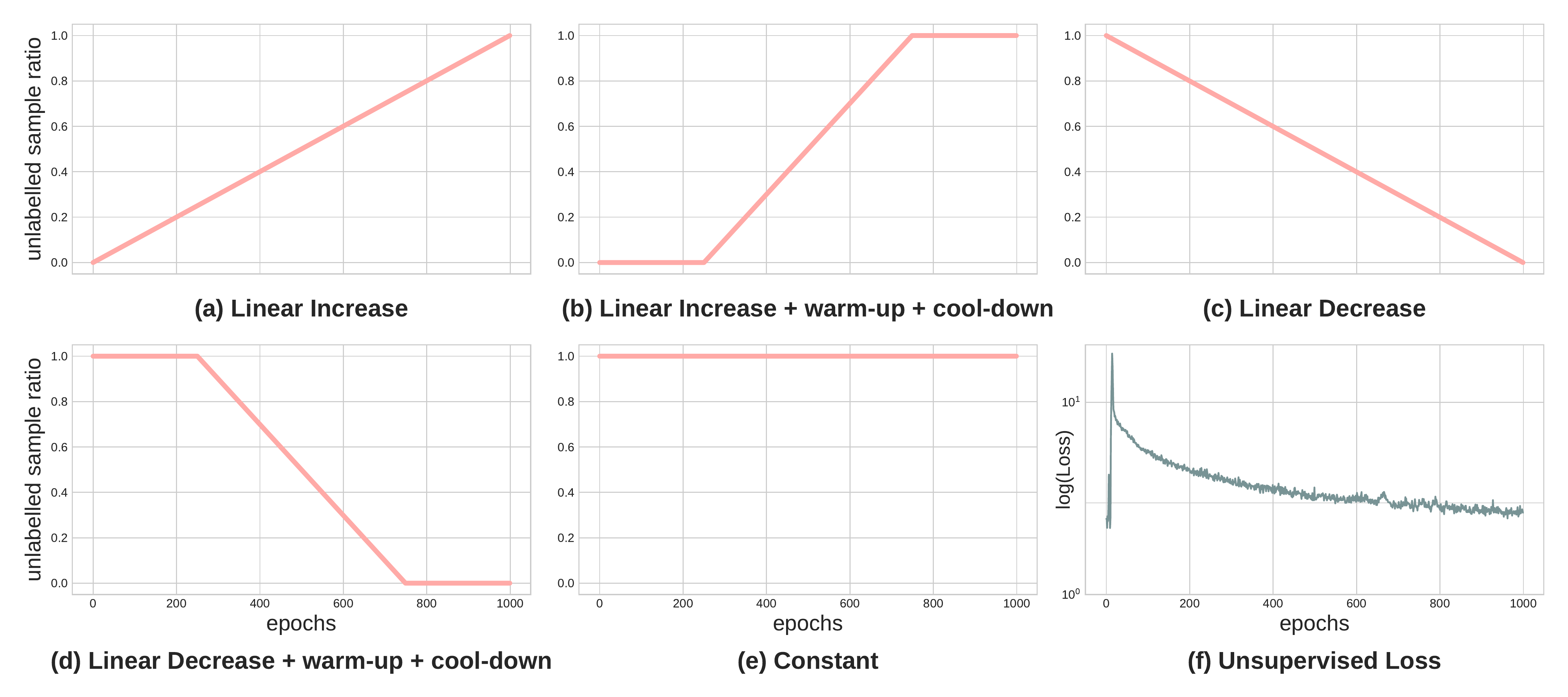}
	\caption{\footnotesize{\textbf{Unlabelled Data Sampling Policies \& Unsupervised Loss Dynamics}. \textbf{(a)} models a linear increase of the unlabelled data ratio from 0 to 1 over epochs, \textbf{(b)} combines {w}arm-{u}p and {c}ool-{d}own phases with linear increase, \textbf{(c)} linear decrease of the unlabelled data ratio, \textbf{(d)} combines linear decrease with {w}arm-{u}p and {c}ool-{d}own phases, \textbf{(e)} keeps a constant ratio, and \textbf{(f)} shows the unsupervised loss $L_\theta^{\prime}$ observed in training.}
	}
	\label{fig:loss}
\end{figure*}

\ \\
\noindent\textbf{Unlabelled Data Sampling Policy} controls the number of unlabelled samples to use in the self-training loop at different time steps. It can be expressed with an additional Bayesian prior on Equation (\ref{equ:lossfunc}), i.e.
\begin{equation}\label{equ:lossprior}
	\footnotesize{
		\begin{split}
			\mathcal{L}_{\pi}(\theta)= & \hat{\mathbb{E}}\left[L_{\theta}\right] \\
			= & \frac{1}{M} \sum_{X_i\in D_X}L_{\theta}\left(X_{i}\right)+\alpha\frac{1}{N\mathbb{E}[\pi]}\sum_{U_j\in D_U}L_{\theta}^{\prime}\left(U_j\right)\pi(U_j),
		\end{split}
	}
\end{equation}
where $\pi(U_j)$ is the probability for using the unsupervised loss of $U_j$ in the self-training stage. For simplicity, we substitute $\alpha\frac{\pi(U_j)}{N\mathbb{E}[\pi]}$ with $p(U_j)$ in Eq. (\ref{equ:lossprior}) to obtain:
\begin{equation}\label{equ:last}
	\footnotesize{
		\begin{split}
			\mathcal{L}_{\pi}(\theta)
			= & \frac{1}{M} \sum_{i=1}^{M} L_{\theta}\left(X_{i}\right)+
			\sum_{j=1}^{N} L_{\theta}^{\prime}\left(U_j\right)p(U_j) \\
			= &\frac{1}{M} \sum_{i=1}^{M} L_{\theta}\left(X_{i}\right)+ \sum_{j=1}^{N} \Big( L_{\theta}^{\prime}\left(U_j\right)p(U_j) -  \hat{\mathbb{E}}[L_\theta^{\prime}]p(U_j) \\
			  &- \hat{\mathbb{E}}[p]L_{\theta}^{\prime}\left(U_j\right)
			  + \hat{\mathbb{E}}[L_\theta^{\prime}]\hat{\mathbb{E}}[p]
		  \Big) + N\hat{\mathbb{E}}[L_\theta^{\prime}]\hat{\mathbb{E}}[p] \\
				=&\frac{1}{M} \sum_{i=1}^{M} L_{\theta}\left(X_{i}\right) +N\hat{\mathbb{E}}[L_\theta^{\prime}]\hat{\mathbb{E}}[p]\\
				 &+\sum_{j=1}^{N} (L_{\theta}^{\prime}\left(U_j\right)-\hat{\mathbb{E}}[L_\theta^{\prime}])(p(U_j)-\hat{\mathbb{E}}[p]) \\
				=&\frac{1}{M} \sum_{i=1}^{M} L_{\theta}\left(X_{i}\right) +N\hat{\mathbb{E}}[L_\theta^{\prime}]\hat{\mathbb{E}}[p] + N\hat{\operatorname{Cov}}[L_\theta^{\prime},p] ~ .\\
			\end{split}
		}
	\end{equation}
	Based on the definition of $\mathcal{L}_{all}(\theta)$ in Eq. (\ref{equ:lossfunc}), Eq. (\ref{equ:last}) can be simplified to:
	\begin{equation}\label{equ:final}
		\footnotesize{
			\begin{split}
				\mathcal{L}_{\pi}(\theta)
				=& \mathcal{L}_{all}(\theta) + N\hat{\operatorname{Cov}}[L_\theta^{\prime},p]
			\end{split}
		}
	\end{equation}
	The goal is to search for the best unlabelled data sampling policy $\pi^*$ that can yield the lowest possible loss for Eq. (\ref{equ:final}), such that:
	\begin{equation}\label{equ:target}
		\footnotesize{
			\begin{split}
				\pi^*=&\underset{\pi}{\arg \min } ~\mathcal{L}_{\pi}(\theta)\\
				=& \underset{\pi}{\arg \min } ~ \mathcal{L}_{all}(\theta) + N\hat{\operatorname{Cov}}[L_\theta^{\prime},p]\\
				=&\underset{\pi}{\arg \min } ~ \hat{\operatorname{Cov}}[L_\theta^{\prime},p]
			\end{split}
		}
	\end{equation}

\noindent Equation (\ref{equ:target}) suggests that if $L_\theta^{\prime}$ and $p$ are negatively correlated then we can arrive at an effective policy $\pi^*$. Given that $p$ is positively correlated with $\pi$, since $\frac{\alpha}{N\mathbb{E}[\pi]}$ is  positive, an effective unlabelled data sampling policy $\pi^*$ should be negatively correlated to $L_\theta^{\prime}$.
	The model gets updated for each iteration, thus one may assume naïvely that $L_\theta^{\prime}(U_{j+1})<L_\theta^{\prime}(U_{j})$, because $L_\theta^{\prime}(U_{j+1})$ is generated after backpropagation of $L_\theta^{\prime}(U_{j})$. In practice, during  training, we also observed such a decrease of $\mathbb{E}[L_\theta^{\prime}]$ as shown in Fig.~\ref{fig:loss}(f).

In summary, considering $\pi^*$ and $L_\theta^{\prime}$ are negatively correlated, and  $L_\theta^{\prime}$ is indeed decreasing over time,  we can conclude that $\pi$ can consequently be obtained via cyclical curriculum learning. Practically, this may be carried out via a gradual increase of unlabelled sample input in the ways shown in Fig.~\ref{fig:loss}(a) or Fig.~\ref{fig:loss}(b), where the latter includes warm-up and cool-down periods. Conceptually, these policies expand learning slowly but steadily towards the unexplored data domain in order to allow for a gradual expansion of high quality model expertise and prevent erratic learning collapse.
	For comparison and to emphasise the importance of this policy choice, we later also experimentally examine other policies depicted in Figs.~\ref{fig:loss} (c), (d), and (e).

\begin{figure*}[ht]
	\centering
	\includegraphics[width=\textwidth]{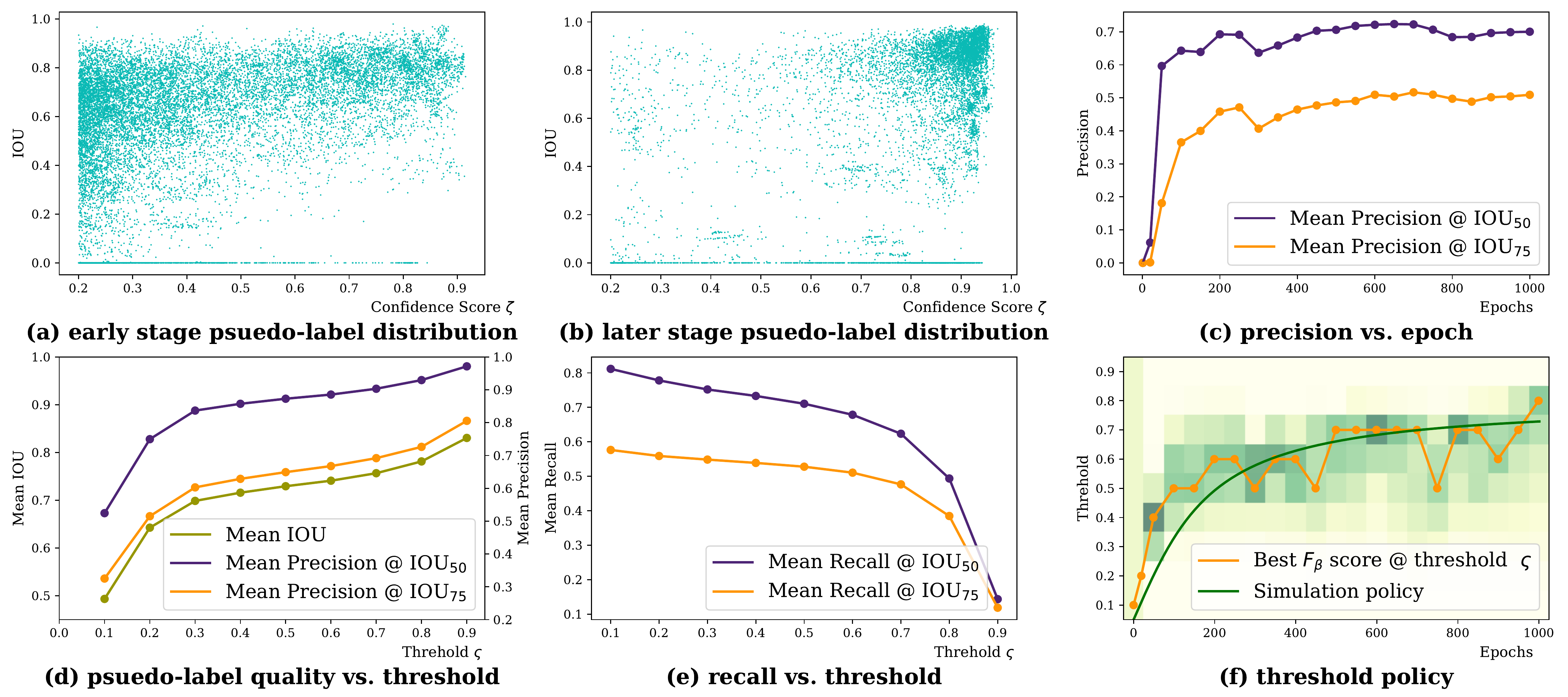}
	\caption{{\footnotesize{\textbf{Pseudo-label Analysis}. We use 70k pseudo-labels generated from the teacher network to conduct this analysis. Note that unlabelled samples do not use ground truth in training, but we use it for this analysis. \textbf{(a)} and \textbf{(b)} show the distributions of each pseudo-label's ground truth IOU against  the confidence score $\zeta$ visualised at the early stage (100$^{th}$ epoch snapshot) and later stage (800$^{th}$ epoch snapshot) of training, respectively. \textbf{(c)} the pseudo-label mean precision at IOU$_{50}$ and IOU$_{75}$ without applying a threshold over the training epochs. \textbf{(d)}  pseudo-label quality against the threshold $\varsigma$ represented by  IOU$_{50}$ and {IOU$_{75}$} precision averaged across epochs, and IoU of real ground truth labels averaged on all epochs. \textbf{(e)} mean recall at IOU$_{50}$ and IOU$_{75}$ for different $\varsigma$ values, \textbf{(f)} heatmap indicating the normalised $F_\beta$ score for $\varsigma_t$ at different epochs -  light to dark colours for low to high  scores, orange shows the best $F_\beta$ score at each epoch, and the dark green plot represents our confidence threshold policy as the $\arctan$ function that approximates the best $F_\beta$ scores.
	}}}
	\label{fig:iouvsthre}
\end{figure*}

\noindent\textbf{Unsupervised Loss Weighting Policy} {is tasked with balancing the weighting between the supervised and  unsupervised losses.}
{The performance of the student detector depends on the quality of the pseudo-labels. Figs. \ref{fig:iouvsthre}(a) and (b) depict pseudo-label distributions captured at an early stage and a late stage of the training, respectively, plotted against the confidence score  $\zeta$. Label confidence and IOU quality clearly increase over training at these snapshot points. The associated @IOU$_{50}$ and @IOU$_{75}$ precision curves in Fig. \ref{fig:iouvsthre}(c) illustrate that the average quality of the pseudo-labels increases over time. We note that recent works \citep{sohn2020simple,xu2021end,tang2021humble} on this topic only applied a fixed weighting to all pseudo-labels throughout the training.
Yet, given this observed gradual change in pseudo-label quality, there is an opportunity to design an adaptive weighting policy that applies smaller unsupervised loss weights for less reliable pseudo-labels in the early training stages and larger weights for more reliable pseudo-labels generated in the later training stages.}

To implement this, we use a curriculum learning approach for the unsupervised loss weighting parameter $\alpha$, which is made subject to an adaptive weighting policy. Theoretically, more optimal policies would keep track of the bounding-box pseudo-label qualities. However, in practice, this is hard to do on the fly due to the unavailability of the ground truth and extensive computational needs. We thus opt for a simple linear increase of $\alpha$ as a first approximation.


\subsection{Pseudo-label Generation}\label{sec:pseudo}
We use two policies to generate reliable pseudo-labels from the teacher's output to promote a virtuous cycle for training.

\noindent\textbf{Confidence Threshold Policy} allows us to examine the reliability of pseudo-labels at different threshold values of $\varsigma$, ranging from 0.1 to 0.9, and averaged across epochs. Our aim is to select an optimised value in order to discard unreliable pseudo-labels most effectively.

Three metrics are applied to assess the quality of the pseudo-labels:  IOU,  Precision @IOU$_{50}$ and {Precision @IOU$_{75}$}. The plots in Fig. \ref{fig:iouvsthre}(d) for {all} measures show that they increase as $\varsigma$ does.
Trivially, the higher the value of $\varsigma$, the higher the probability of obtaining more reliable pseudo-labels. So for highest quality one could select $\varsigma=0.9$. However, as a consequence the recall rate is significantly suppressed, with both mean @IOU$_{50}$ and @IOU$_{75}$ recalls of course negatively correlated to $\varsigma$ (see Fig. \ref{fig:iouvsthre}(e)). For example, when $\varsigma=0.9$, mean precision reaches approx. $95\%$, while the mean recall drops to approx. $15\%$.

\begin{figure*}[ht]
	\centering
	\includegraphics[width=0.99\linewidth]{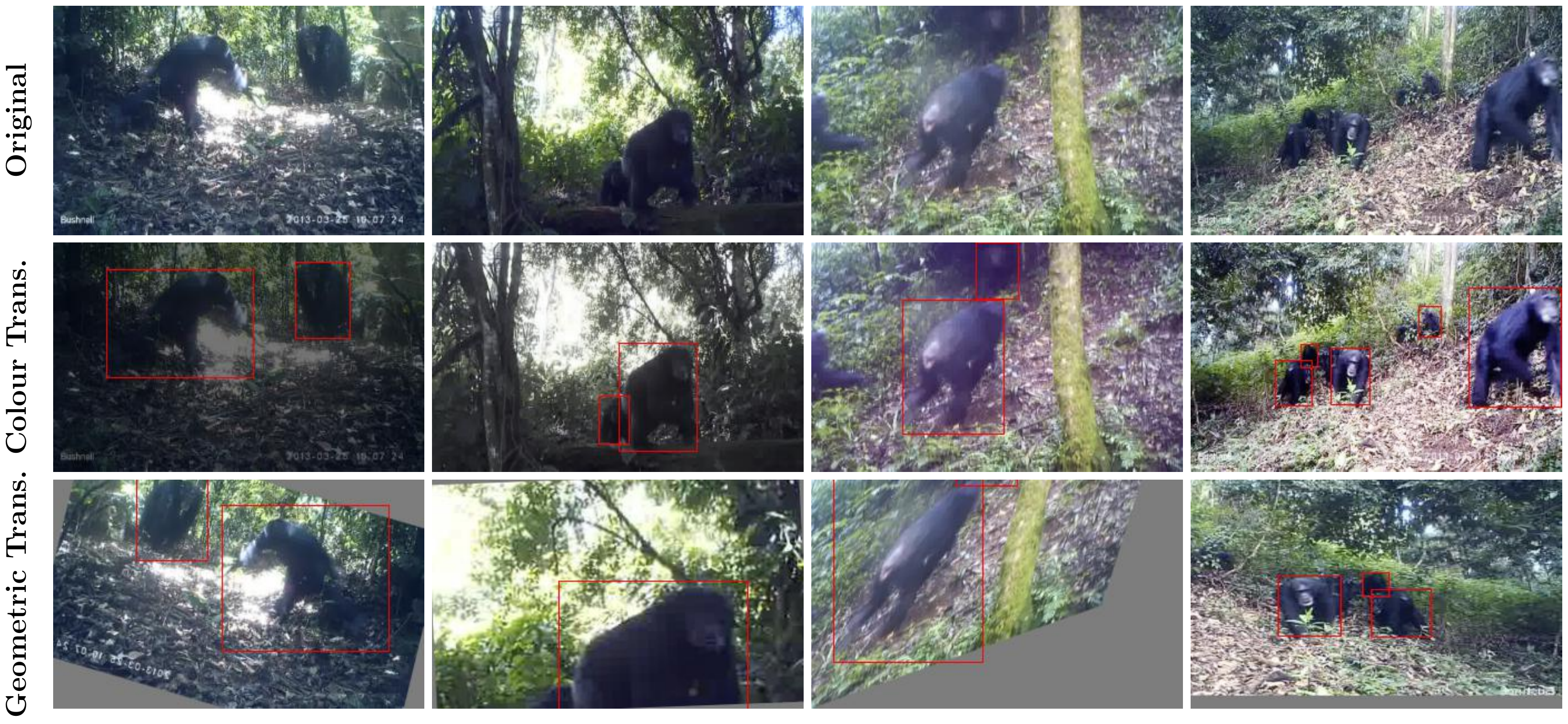}
	\caption{\footnotesize{{\textbf{Augmentation Strategy}. Visualisation of colour augmentation and geometric augmentation examples used in the experiments. Augmentations are selected such that the results reflect the variance found across different camera and acquisition settings commonly seen in the dataset. (best viewed under zoom)
	}}}
	\label{fig:aug}
\end{figure*}

To address this issue, our dynamic confidence threshold policy increases the quality of pseudo-labels by controlling false negatives explicitly and thereby balancing precision and recall. We apply the $F_{\beta}$ score which is the weighted harmonic mean of precision and recall, with $\beta=0.5$ to allow the $F_{\beta}$ score assign more weight to the precision than the recall on the basis that  the false positives have more negative impact than the false negatives in our pipeline. Compared to  anchor-based detectors which assign missed bounding boxes as negatives (non-object class), such as  {\cite{ren2015faster,he2017mask,lin2017focal}}, the DETR family of methods do not. Their bipartite matching stage only matches the predictions with the ground truth, thus any missed bounding boxes are ignored and there are no penalties for this in training. Further, bounding box-aware crops are applied in the augmentation stage, 
thus false negative areas could be wiped out in the image. For example,  see the chimp on the right side of the first image in the last column in Fig. \ref{fig:aug}, which if undetected by the teacher, it will disappear after a geometric transformation, as shown in the last image in the last column.

After fixing $\beta$, which may be changed for different application scenarios, the goal of our confidence threshold policy is to search for the threshold $\varsigma$ at time step $t$ that can maximise $F_{\beta}$, i.e.
\begin{equation}\label{equ:threshold}
	\footnotesize{
		\begin{array}{c}
			\varsigma^*_{t}=\underset{\varsigma_t}{\arg \max}~ F_{\beta} (\mathcal{P}_t , \mathcal{R}_t | \varsigma_t),\\
		\end{array}
	}
\end{equation}
where $\mathcal{P}_t$ and $\mathcal{R}_t$ represent the precision and recall rates at time step $t$, determined by threshold $\varsigma_t$. The heatmap in Fig. \ref{fig:iouvsthre}(f) shows the normalised $F_\beta$ score for each time step (darker colour means higher value), with the best $F_\beta$ at threshold $\varsigma_t$ shown in the orange plot. We use the approximate fit of the $\arctan$ function (dark green plot) to represent our confidence threshold policy.

\noindent\textbf{Data Augmentation Policy} ensures consistent augmentation of unlabelled data under the pseudo-labels produced by the teacher.

This is an indispensable element in semi-supervised and self-supervised methods. Self-supervised methods, such as DINO \citep{caron2021emerging} and BYOL \citep{grill2020bootstrap} minimise different views of data generated by data augmentation. Recently, semi-supervised methods such as FixMatch \citep{sohn2020fixmatch} and STAC \citep{sohn2020simple}, use augmentation-driven consistency regularisation for classification and detection.

\begin{algorithm*}[ht]
\caption{Semi-supervised Training of Great Ape Detector using all Policies}\label{alg}
\begin{algorithmic}[1]
\footnotesize
\State \textbf{Require:} {$D_X,D_{U} $} \Comment{ labelled and unlabelled data}
\State \textbf{Require:} {$\mathfrak{S}_\theta, \mathfrak{T}_{\theta^{\prime}}$} \Comment{ student and teacher models}
\State \textbf{Require:} {$\Pi$} \Comment{ curriculum learning strategy}
\State {$\mathfrak{S}_\theta$} $\gets$ initialisation {$\mathfrak{T}_{\theta^{\prime}}$} $\gets$ initialisation \Comment{ initialise student and teacher models}
\Do
\State $\pi_t, \varsigma_t, \alpha_t, m_t, \mathcal{A}$ $\gets$ $\Pi$(t) \Comment{ instantiate five policies at each time step}
\State $X_i$, $[C_i, B_i]$ $\gets$ \textit{mini-batch}($D_X$) \Comment{ sample labelled mini-batch data}
\State $U_j$ $\gets$ \textit{mini-batch}($D_U$)  \Comment{ sample unlabelled mini-batch data}
\State $U_j$ $\gets$ $\pi_t(U_j)$ \Comment{ apply unlabelled data sampling policy}
\State $\mathcal{A}_s,\mathcal{A}_w$ $\gets$ $\mathcal{A}$ \Comment{ sample strong and weak augumentations}
\State $[C_j,B_j], \zeta_j$ $\gets$  $\mathfrak{D}_{\theta^{\prime}}(\mathcal{A}_w(U_j))$ \Comment{ generate pseudo-labels by teacher}
\State $\stackrel\frown{U_j},[\stackrel\frown{C_j},\stackrel\frown{B_j}]$ $\gets$ $\mathbbm{1}(U_j,[C_j,B_j] ~ | ~ \zeta_j>\varsigma^{t})$
\Comment{ apply confidence threshold policy}
\State $\mathcal{L}$ $\gets$ \textit{Loss}$\big(\mathfrak{D}_\theta(\mathcal{A}_w(X_i)), \mathcal{A}_w([C_i,B_i])\big)$ \Comment{ get supervised loss with ground truth}
\State $\mathcal{L^{\prime}}$ $\gets$ \textit{Loss}$\big(\mathfrak{D}_{\theta}(\mathcal{A}_s(\stackrel\frown{U_j})), \mathcal{A}_s([\stackrel\frown{C_j},\stackrel\frown{B_j}])\big)$ \Comment{ get unsupervised loss with pseudo-labels}
\State $\mathcal{L}_{all}$ $\gets$ $\mathcal{L}+\alpha_t\mathcal{L^{\prime}}$ \Comment{ apply unsupervised loss weighting policy}
\State $\Delta\theta$ $\gets$ $-\nabla_{\mathcal{L}_{all}}\theta$ \Comment{ backpropagate the overall loss}
\State $\theta$ $\gets$ $\theta+\Delta\theta$ \Comment{ undate student networks by gradient}
\State $\theta^{\prime}$ $\gets$ $m_t\theta^{\prime}+(1-m_t)\theta$\Comment{ undate teacher with teacher momentum policy }
\State $t$ $\gets$ $t+1$ \Comment{ next time step and repeat}
\doUntil{$\mathcal{L}_{all}$ converge}
\State \textbf{end}
\end{algorithmic}
\end{algorithm*}

Following STAC, we explore different variants of transformations on the Extended PanAfrican Dataset as our augmentation policy $\mathcal{A}$. We apply transformation operations in sequence as follows: first, we randomly apply bounding-box-aware crop and resize on the image, and then we apply a randomly-selected geometric transformation, followed by a random transformation on the colour statistics of the image (see code for all details).

Finally, for strong augmentation $\mathcal{A}_s$, we apply random erase \citep{zhong2020random} or cutout \citep{devries2017improved} at multiple random locations of the whole image. For a weak augmentation $\mathcal{A}_w$, we just decrease the intensity for each transformation. Some examples are shown in Fig. \ref{fig:aug} to illustrate the augmentation process.

\subsection{Teacher Update}
\textbf{Teacher Momentum Policy} controls the update speed of the teacher model and is encapsulated in the momentum coefficient $m$. In Fig. \ref{fig:iouvsthre}(c), we can see the pseudo-label precision rate increases steeply in the early training stages, but slowly in the later training stages. Given that learning happens faster in the early stage, it motivates us to design a dynamic momentum policy which takes this fact onboard and stabilises teacher updates. Eq. (\ref{equ:momentum}) suggests that a lower momentum coefficient allows faster updates of the teacher model. To match the learning speed of the model at different time steps, we use a lower momentum coefficient $m$ at the early stages and gradually increase it with time. In practice, we use a cosine increase of $m$ in our pipeline which has also been explored in DINO \citep{caron2021emerging}.
In experiments, we find that this dynamic momentum policy leads to consistently better performance than a constant one (see Table~\ref{tab:ablation}).

\subsection{Combined Policy Application}
All five policies are implemented in unison as a dynamic curriculum learning strategy $\Pi$ for our wildlife detection pipeline, comprising the unlabelled data sampling policy $\pi_t=\Pi_\pi(t)$, the unsupervised loss weighting policy $\alpha_t=\Pi_\alpha(t)$, the confidence threshold policy $\varsigma_t=\Pi_\varsigma(t)$, the data augmentation policy $\mathcal{A}=\Pi_{\mathcal{A}}(t)$  and the teacher momentum policy $m_t=\Pi_m(t)$. Algorithm~\ref{alg} illustrates this curriculum learning strategy $\Pi=\{\pi_t,\alpha_t,\varsigma_t,\mathcal{A},m_t\}$ in its complete form.


\section{Experiments}
\label{lab:experiments}
{\bf Datasets -- }{We test our method on the Extended PanAfrican Dataset from the PanAf programme \citep{mpi} which contains camera-trap footage captured in natural Great Ape habitats in central Africa.
There are two major species of Great Apes in the dataset, gorillas and chimpanzees. The archive footage contains around 20K videos adding up to around 600 hours. We use a subset of 5219 videos, with 500 videos (totalling over 180K frames) manually annotated with per frame great ape location bounding boxes, species and further categories~\citep{yang2019great,sakib2020visual}. This labelled data is split into \texttt{trainset}, \texttt{valset}, \texttt{testset} at a ratio of $80\%, 5\%, 15\%$ respectively. All labels and metadata are fully published~\citep{yang2019great} and source videos may be obtained as detailed in the Acknowledgements.}

\begin{table*}[!ht]
\centering
\small
\begin{tabular}{lcc|ccc}
	\toprule
	\textbf{Method}               & \textbf{Labelled Ratio} & \textbf{Setting} & mAP              & mAP$_{50}$                 & mAP$_{75}$                 \\
\midrule \midrule
Supervised baseline                   & $10\%$                  & PLD              & $32.17\pm0.70$             & $75.57\pm1.24$             & $21.40\pm2.45$             \\
STAC \citep{sohn2020simple}           & $10\%$                  & PLD              & $38.04\pm3.88$             & $73.31\pm7.01$             & $35.34\pm2.31$             \\
SoftTeacher$^*$ \citep{xu2021end}     & $10\%$                  & PLD              & $39.37\pm7.97$             & $63.03\pm11.42$            & $44.50\pm9.72$             \\
Ubteacher$^*$ \citep{liu2021unbiased} & $10\%$                  & PLD              & $\underline{44.03}\pm0.26$ & $\underline{76.69}\pm2.15$ & $\underline{47.25}\pm1.21$ \\
Ours                                  & $10\%$                  & PLD              & $\textbf{45.96}\pm{2.97}$  & $\textbf{78.10}\pm{6.14}$  & $\textbf{47.67}\pm3.12$    \\
\midrule
Supervised baseline                   & $20\%$                  & PLD              & $46.93\pm1.30$             & $86.47\pm0.74$             & $46.00\pm2.42$             \\
STAC                                  & $20\%$                  & PLD              & $51.35\pm2.39$             & $83.71\pm2.24$             & $56.58\pm4.12$             \\
SoftTeacher$^*$                       & $20\%$                  & PLD              & $50.87\pm2.99$             & $79.57\pm6.29$             & $58.67\pm2.89$             \\
UbTeacher$^*$                         & $20\%$                  & PLD              & $\underline{55.78}\pm0.45$ & $\underline{88.07}\pm1.88$ & $\underline{63.02}\pm0.67$ \\
Ours                                  & $20\%$                  & PLD              & $\textbf{59.01}\pm1.57$    & $\textbf{89.23}\pm0.98$    & $\textbf{66.95}\pm2.45$    \\
\midrule
Supervised baseline                   & $50\%$                  & PLD              & $59.50\pm1.40$             & $\underline{92.37}\pm0.92$ & $65.47\pm2.04$             \\
STAC                                  & $50\%$                  & PLD              & $59.93\pm1.21$             & $92.35\pm0.65$             & $67.40\pm2.10$             \\
SoftTeacher$^*$                       & $50\%$                  & PLD              & $60.47\pm3.58$             & $86.93\pm4.23$             & $69.63\pm3.35$             \\
UbTeacher$^*$                         & $50\%$                  & PLD              & $\underline{61.66}\pm1.73$ & $91.79\pm1.45$             & $\textbf{72.71}\pm1.42$    \\
Ours                                  & $50\%$                  & PLD              & $\textbf{63.39}\pm1.34$    & $\textbf{92.96}\pm0.68$    & $\underline{70.00}\pm3.45$ \\
\midrule \midrule
Supervised baseline                   & $100\%$                 & FLD              & 65.53                      & \underline{95.28}                      & 74.52                      \\
STAC                                  & $100\%$                 & FLD              & 46.98                      & 80.76                      & 50.61                      \\
SoftTeacher$^*$                       & $100\%$                 & FLD              & \textbf{70.70}             & 94.90                      & \textbf{81.90}             \\
UbTeacher$^*$                         & $100\%$                 & FLD              & 66.45                      & 94.13                      & \underline{79.35}          \\
Ours                                  & $100\%$                 & FLD              & \underline{67.64}          & \textbf{95.87}             & 76.81                      \\

\bottomrule
\end{tabular}
\caption{\footnotesize{{\textbf{Results and Detailed Comparative Evaluation on the Extended PanAfrican Dataset  --} Mean and standard deviation on test set portion evaluated over 3 data folds for $10\%$, $20\%$ and $50\%$ Labelled Ratio are reported. Supervised baseline refers to the same model trained on the labelled data only. Other state-of-the-art methods are re-evaluated on the dataset based on their  publicly available codebase. We evaluate the methods with PLD and FLD settings which represent the Partially Labelled Data and Fully Labelled Data paradigms. PLD evaluation in particular was performed at scale using the Labelled Ratio portion of 500 labelled videos (i.e. $\sim$180k annotated frames) as labelled input and adding remaining videos plus $\sim$5000 additional unlabelled videos (i.e. $\sim$1.8M frames) of the same domain for unlabelled input. Note that $*$ indicates the code's data loader was changed for this dataset.}}
\label{tab:sota}
}
\end{table*}
Following standard evaluation protocols as used in~\citep{sohn2020simple,xu2021end,zhou2021instant,liu2021unbiased}, we utilise the Extended PanAfrican Dataset for system training and benchmarking under two general paradigms:

\begin{enumerate}
	\item\textbf{Partially Labelled Data (PLD)}. In this setting, either $10\%$, $20\%$, or $50\%$  of the annotated \texttt{trainset} data are sampled as labelled training data, and the complete remainder of all data is used as unlabelled data. For each quantity, we create 3 different data folds and report the performance on \texttt{testset} with mean average precision (mAP) as the evaluation metric.
		\\ \
	\item \textbf{Fully Labelled Data (FLD)}. In this setting, the whole annotated \texttt{trainset} is utilised as the labelled training data and only the remaining $\sim$5K unlabelled videos, totalling $\sim$1.8M frames, are used as additional unlabelled data.
\end{enumerate}

In addition, we investigate two other animal datasets under sparse labelling settings - Bees\footnote{Available at \url{https://lila.science/datasets/boxes-on-bees-and-pollen}} and Snapshot Serengeti \citep{swanson2015snapshot} - to explore system effectiveness under sparse labelling regimes further across the domain of animal visuals.
We also present results on the MS-COCO  \citep{lin2014microsoft} and {PASCAL-VOC \citep{everingham2010pascal} datasets} to explore wider applicability of the introduced concepts to mainstream object detection.

\ \\
\noindent{\bf Implementation Details --}
We use a Deformable DETR architecture with a ResNet-50 backbone as our default detection model~(see Fig.~\ref{fig:overview}) for evaluating the effectiveness of our method. The transformer decoder and encoder are randomly initialised and the ImageNet pre-trained ResNet-50 weights from SWAV \citep{caron2020unsupervised} are used as initial parameters for our backbone. The student model is trained with the AdamW optimizer~\citep{loshchilov2018fixing} with a weight decay of 0.0004 and a batch size of 64, distributed over 4 GPUs. We follow \cite{caron2021emerging} using a linear scale rule of $lr = 0.0005\times batchsize /64$ and apply a slightly lower learning rate of $0.1\times lr$ for the backbone.

We use randomly sampled frames for each video at each epoch with frequency $\omega=10$, and the frames are rescaled so that the smaller axis of the frame is in range $[320,480]$. The PLD model is trained for 1000 epochs with the first quarter as the warm-up phase and the last quarter as the cool-down phase (Fig. \ref{fig:loss}(a)), and $lr$ decreases to $5e-5$ at the $800^{th}$ epoch. The momentum $m$ for updating the teacher follows a cosine schedule from 0.998 to 0.9998. Since the amount of training data for the partially labelled data setting and the fully labelled data setting is quite different, training parameters vary slightly from that for FLD\footnotemark.
\footnotetext{{For FLD, we use total epochs=1100 with the first 500 as warmup, the last 100 as cooldown and  $lr$ decreases at the $1000^{th}$ epoch. The unlabelled ratio is bounded at $10$ in minibatch. Linear increase $\alpha$ is $0.3\to1$ and arctan increase is $\varsigma$ $0.3\to0.6$. }}\newpage

\begin{figure}[th]
	\centering
	\includegraphics[width=\linewidth]{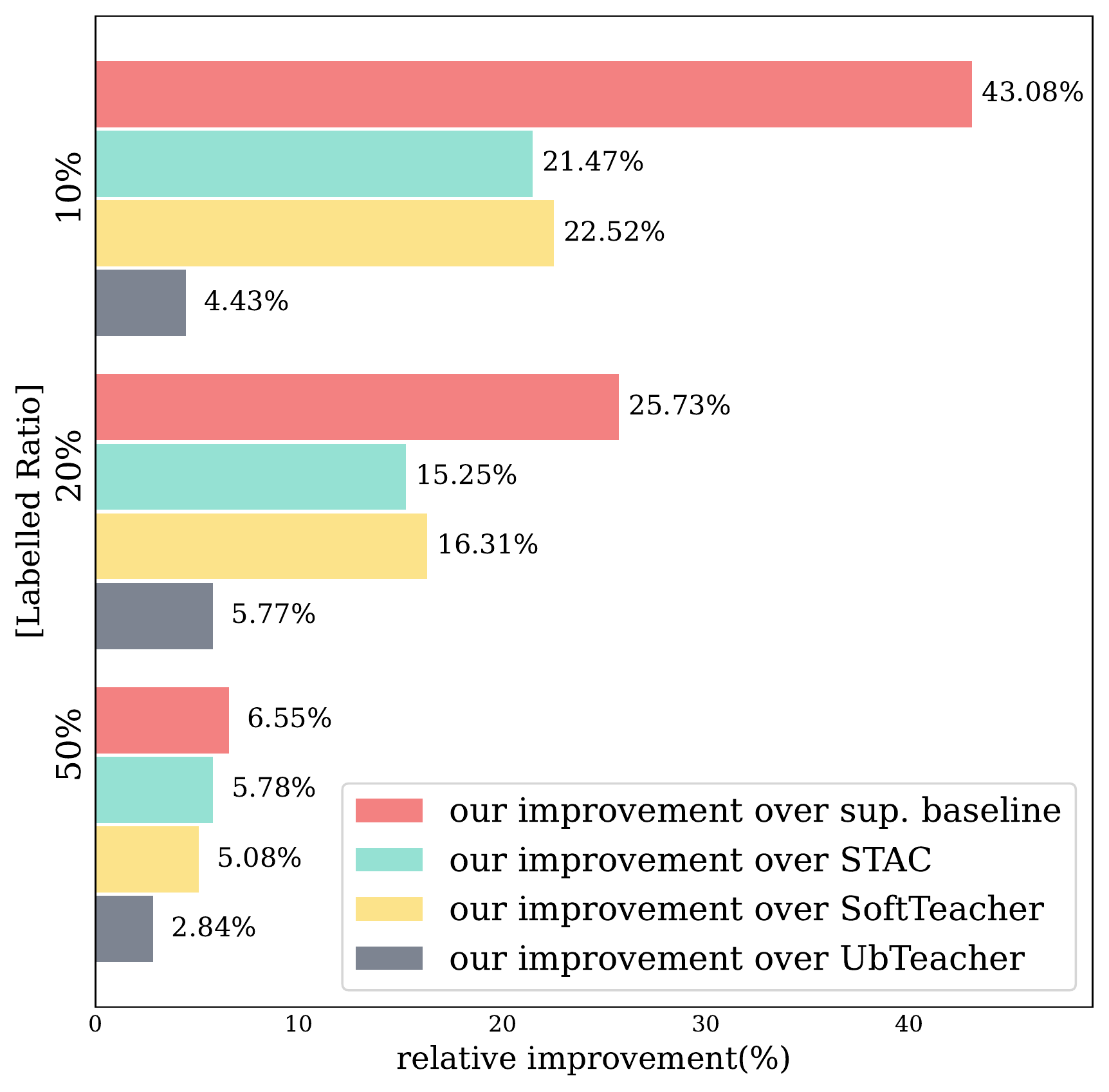}
	\caption{\footnotesize{{\textbf{Relative Improvement Comparisons.} Relative improvement of mAP for our method over the supervised baseline, STAC, SoftTeacher, and UbTeacher across various PLD settings. We find our method shows particularly strong performance in lower annotation ratio regimes typical for many wildlife data settings.}}}
	\label{fig:comparision}
\end{figure}

\noindent{\bf Comparative Evaluation -- } We first evaluate our method for the PLD and FLD settings against a supervised baseline and state-of-the-art works STAC \citep{sohn2020simple}, SoftTeacher \citep{xu2021end} and {UbTeacher} \citep{liu2021unbiased} at various ratios of labelled data. Table~\ref{tab:sota} summarises the results.

\begin{figure*}[ht]
	\centering
	\includegraphics[width=\linewidth,height=190pt]{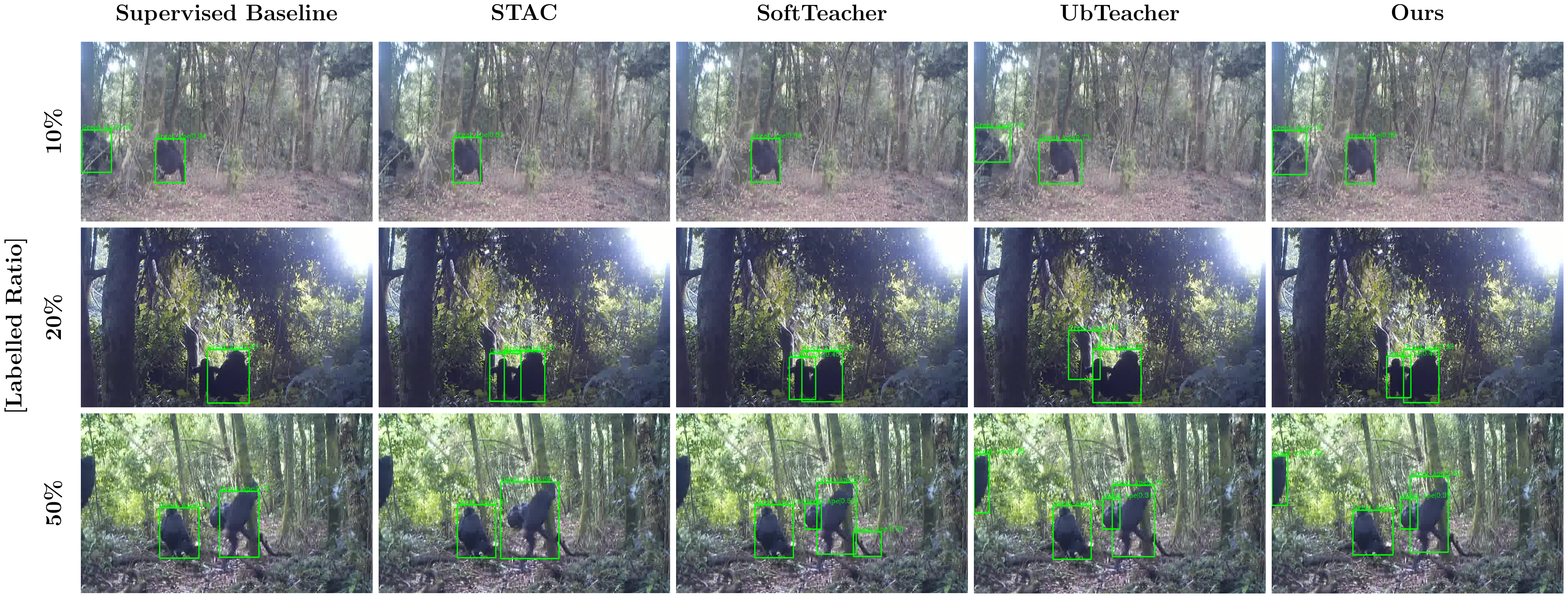}
	\caption{\footnotesize{\textbf{Qualitative Detection Examples}. We compare our method with other state-of-the-art approaches tested and the supervised baseline under the PLD setting with Labelled Ratios of $10\%$, $20\%$, $50\%$. Note examples where our proposed method reliably detects partly occluded apes and ignores tree structures which distract some of the other models. (best viewed under zoom)
	}}
	\label{fig:compare}
\end{figure*}

\begin{figure*}[ht]
	\centering
	\includegraphics[width=\linewidth,height=190pt]{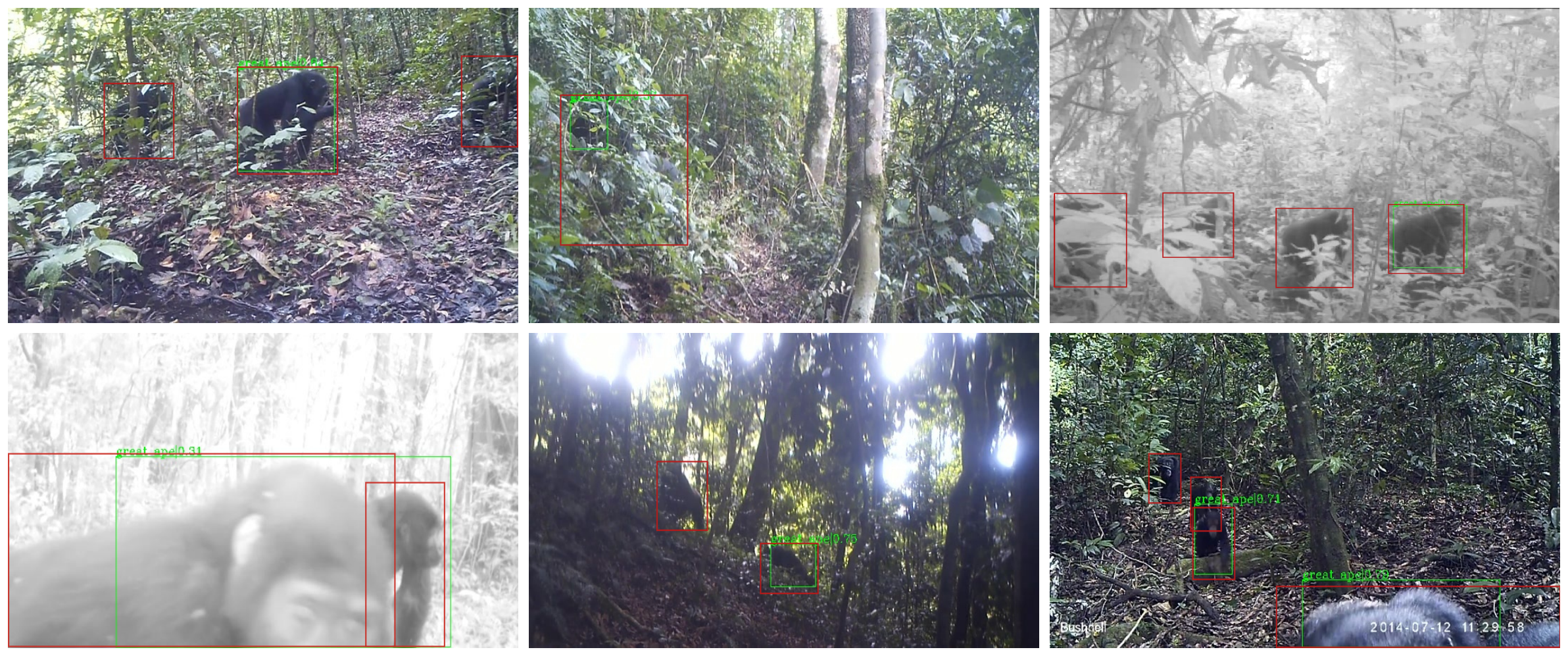}
	\caption{\footnotesize{\textbf{Examples of Failure Cases}. Visualised are failure cases under the $10\%$ PLD setting. Ground-truth labels are annotated in red, and our detection results are shown in green. Note that partial occlusions form one hard-to-learn aspect given sparse label availability for training (best viewed under zoom).}}
	\label{fig:show_neg}
\end{figure*}

Our proposed method shows significant performance improvements under almost all test settings. For example, in the mAP column, we outperform the supervised baseline by $13.79\%$, $12.08\%$, $3.89\%$, STAC by $7.92\%$, $7.66\%$, $3.46\%$, SoftTeacher by $6.59\%$, $8.14\%$, $2.92\%$ and UbTeacher by $1.93\%$, $3.23\%$, $1.73\%$ when $10\%$, $20\%$, $50\%$ of labelled data are provided, respectively. We find that our method works better than others particularly when the provided labelled data is small as illustrated in Fig. \ref{fig:comparision}. We note again that such a setting is particularly common in wildlife applications where camera trap archives are large and accurate annotation ratios are very small. We can see that competitor methods also show sizeable improvements over the supervised baseline for smaller splits, indicating unsurprisingly that extra unlabelled data has particularly high value when very little labelling is available in the first place. However, note that the performance gap between the proposed method and other approaches is also particularly large in exactly this setting, confirming the specific applicability of our enhanced dynamics for curriculum learning in low labelling ratio settings. Qualitative results across all methods are exemplified and discussed in Fig.~\ref{fig:compare}. This is complemented by visualisations of some failure cases in Fig.~\ref{fig:show_neg}.

\begin{table*}[!ht]
\begin{minipage}[t]{0.32\textwidth}
\begin{center}
\textbf{\footnotesize\textit{(a)}}\\
\small
\begin{tabular}[t]{c|c}
    \textbf{Sample Policy $\pi$}                              & \textbf{mAP}   \\
                                                                              \\ \hline
    Linear Increase                                          & \underline{61.62}          \\
    Linear Increase$^{\dagger*}$                             & \textbf{64.73} \\
    Linear Decrease                                          & 53.47          \\
    Linear Decrease$^\dagger$                                & 59.99          \\
    Constant                                                 & 60.46          \\
	\end{tabular}
\end{center}
\end{minipage}
\begin{minipage}[t]{0.32\textwidth}
\begin{center}
\textbf{\footnotesize\textit{(b)}}\\
\small
\begin{tabular}[t]{c|c}
\textbf{Loss Weight Policy $\alpha$} & \textbf{mAP}   \\
                                                                             \\ \hline
Constant $0.1$                  & 62.18          \\
Constant $0.5$                  & \underline{63.98}          \\
Constant $1.0$               &  63.21         \\
Constant $2.0$                  & 54.68          \\
Linear $0.1\to 1^*$             & \textbf{64.73} \\
Linear $0.1\to 2$               & 61.17          \\
	\end{tabular}
\end{center}
\end{minipage}
\begin{minipage}[t]{0.32\textwidth}
\begin{center}
\textbf{\footnotesize\textit{(c)}}\\
\small
	\begin{tabular}[t]{c|c}
\textbf{Conf. Threshold Policy $\varsigma$}                                & \textbf{mAP}   \\   \\ \hline
	Constant $0.05$                                         & 51.73          \\
	Constant $0.6$						& 62.62          \\
	Constant $0.9$                                          & 58.61          \\
	Linear $0.3\to 0.5$                                     & 62.38          \\
	Linear $0.1\to 0.6$                                     & \underline{64.10}          \\
	$\arctan$ $0.1\to 0.6^*$                                & \textbf{64.73} \\
	\end{tabular}
	\end{center}
\end{minipage}\\ \ \\
\begin{minipage}[t]{0.32\textwidth}
\begin{center}
\textbf{\footnotesize\textit{(d)}}\\
\small
	\begin{tabular}[t]{c|c}
\textbf{Augmentation Policy $\mathcal{A}$}                                          & \textbf{mAP}   \\
                                                                             \\ \hline
  No Augmentation                                                                    & \underline{55.41}          \\
  Augmentation with $\mathcal{A}_w$, $\mathcal{A}_s^*$                                    & \textbf{64.73} \\
	\end{tabular}
	\end{center}
\end{minipage}
\begin{minipage}[t]{0.32\textwidth}
\begin{center}
\textbf{\footnotesize\textit{(e)}}\\
\small
\begin{tabular}[t]{c|c}
\textbf{Momentum Policy $m$} & \textbf{mAP}    \\
                                               \\ \hline
Constant $0.999$                                    & \underline{60.07}          \\
Constant $0.998$				    & 59.19          \\
Constant $0.9998$                                   & 57.60          \\
$\cos$ $0.998\to0.9998^*$                           & \textbf{64.73} \\
\end{tabular}
\end{center}
\end{minipage}
\begin{minipage}[t]{0.32\textwidth}
\begin{center}
\textbf{\footnotesize\textit{(f)}}\\
\small
\begin{tabular}[t]{c|c}
\textbf{Initialisation}                          & \textbf{mAP}   \\
                                                                  \\ \hline
 Random Init.                                    & 0.17           \\
 SWAV Init.$^*$                                  & \textbf{64.73} \\
 supervised Init.                                & \underline{61.55}          \\
\end{tabular}
\end{center}
\end{minipage}
\caption{\footnotesize{\textbf{Ablation Studies of all Learning Policies.} The effectiveness of the introduced policies are verified via ablation. All studies are conducted for the \texttt{fold1} split at $50\%$ Labelled Ratio in the PLD setting. The symbol $^*$ represents the default settings which we use in the system and the $^{\dagger}$ symbol  denotes the presence of warm-up and cool-down phases. Only one policy is varied for each study to isolate the effect. `Constant' represents `no curriculum' learning where the related variable or unlabelled sample pool stays constant.}}
\label{tab:ablation}
\end{table*}

\section{Ablation Studies}\label{sec:abu}

In this section, we evaluate our key contributions by examining the importance of each policy. We use the \texttt{fold1} split of the $50\%$ PLD setting as the base for the conducted ablations.

\ \\
\noindent\textbf{Unlabelled Data Sample Policy --} The motivation of our unlabelled data sample policy (Eq. \ref{equ:target}) is to make sure that if the unlabelled samples loss $L_\theta^{\prime}$ and unlabelled data sampling policy $\pi$ are negatively correlated, the minimum possible loss can be achieved. Based on this hypothesis, we designed five experiments for five different possible characteristics for policy $\pi$: (i) linear increase of the number of unlabelled samples (Curriculum learning) used in training over the training process (depicted in Fig. \ref{fig:loss}(a)),
(ii) linear increase but with warm-up and cool-down phases at the beginning and at the end respectively (Fig. \ref{fig:loss}(b)), (iii) start with all unlabelled samples but linearly decrease the number of unlabelled samples throughout training (Fig. \ref{fig:loss}(c)), (iv) The opposite of (ii) (Fig. \ref{fig:loss}(d)), (v) using all unlabelled samples constantly throughout training (Fig. \ref{fig:loss}(e)).

The results in Table \ref{tab:ablation}(a) show that a gradual increase of the number of unlabelled samples  during the self-training phase can gain around $1.16\%$ mAP compared with constantly using all the unlabelled samples. The best performance however is achieved by introducing a warm-up and a cool-down phase at $64.3\%$ mAP. This ablation experiment demonstrates that both the choice of 'phasing in' unlabelled data underpinned by our theoretical discussion in Section~\ref{studentintervention} have a positive measurable effect on learning performance.

\ \\
\noindent\textbf{Unsupervised Loss Weight Policy --} The results in Table \ref{tab:ablation}(b) demonstrate the effects of the unsupervised loss weight policy. We find that setting the unsupervised loss weight $\alpha$ is a challenge since both a large loss weight and a small loss weight can harm the performance. A $9.30\%$ mAP drop occurs when $\alpha=2.0$ compared to $\alpha=0.5$. We argue that constantly applying a large $\alpha$ would harm the training at the beginning because $\alpha$ would assign a large weight for the loss produced by the unreliable pseudo-labels in the early stages which would mislead the model, causing it to get caught up in vicious training cycles. In contrast, applying our dynamic weighting approach, the performance reaches $64.73\%$, which is $0.75\%$ better than a constant $\alpha=0.5$, $2.55\%$ better than $\alpha=0.1$  and about $10\%$ better than $\alpha=2$.
As discussed in Sec. \ref{studentintervention}, while  our linear increase performs best, it is a naïve approach since the best global policy is difficult and computationally costly to find across the space of monotonously growing functions. Further exploration of this policy is subject to future work.

\ \\
\noindent\textbf{Confidence Threshold Policy -- } Table \ref{tab:ablation}(c) displays the effects of different approaches for the confidence threshold policy.
Both low and high thresholds cause significant performance degradation, at both low and high thresholds, e.g. $\varsigma=0.05$ and  $\varsigma=0.9$, respectively, with lower thresholds being worse.
This suggests that false positive pseudo-labels (appearing at low thresholds) have more negative impact than the false negative pseudo-labels (that appear at higher thresholds).
As noted in Section \ref{sec:pseudo}, this motivated us to use a new weighted metric $F_\beta$ to assess the best choice of threshold for this policy. Applying the $\arctan$ increasing $\varsigma$ approach, we see a significant increase in performance.

For comparison, we conducted a linear increase approach from 0.1 to 0.6, which takes similar strides, although $\arctan$ increases more aggressively in the early stages and achieves a  slightly better outcome.
\newpage
\noindent\textbf{Augmentation Policy -- } A simple ablation is performed on the proposed augmentation policy, i.e.  $\mathcal{A}_w$ and $\mathcal{A}_s$ augmentations versus no augmentation. Results show a significant improvement by $9.32\%$ as seen in Table \ref{tab:ablation}(d).

\ \\
\noindent\textbf{Teacher Momentum Policy --} We compare a static momentum coefficient approach  with a dynamic momentum approach for $m$ which is used to update the teacher network. As shown in Table \ref{tab:ablation}(e),  both settings have a similar expected value but the dynamic momentum policy improves the performance significantly by $4.66\%$.

\ \\
\noindent\textbf{Initialisation --}
The initial status of the student model is crucial since it can affect training direction from the start towards an effective virtuous or catastrophic vicious cycle of learning. We see in Table \ref{tab:ablation}(f) that a random initialisation of the model can lead to such catastrophic failure in training, while a SWAV-based self-supervised initialisation outperforms a supervised one.

\begin{table*}[ht]
\centering
\begin{tabular}{l|c|c|c|c}
\toprule
\textbf{Method}                          & \textbf{Venue} & \textbf{1\% PLD}           & \textbf{5\% PLD}           & \textbf{10\% PLD} \\
\midrule \midrule
STAC \citep{sohn2020simple}              & Arxiv'20       & 13.97$\pm$0.35             & 24.38$\pm$0.12             & 28.64$\pm$0.21             \\
Instant-Teaching \citep{zhou2021instant} & CVPR'21        & 18.05$\pm$0.15             & 26.75$\pm$0.05             & 30.40$\pm$0.05             \\
Humble-Teacher \citep{tang2021humble}    & CVPR'21        & 16.96$\pm$0.38             & 27.70$\pm$0.15             & 31.61$\pm$0.28             \\
Unbiased-Teacher \citep{liu2021unbiased} & ICLR'21      & \underline{20.75}$\pm$0.12 & 28.27$\pm$0.11             & 31.50$\pm$0.10             \\
Soft-Teacher \citep{xu2021end}           & ICCV'21      & 20.46$\pm$0.39             & \textbf{30.74}$\pm$0.08    & \underline{34.04}$\pm$0.14 \\
DETReg   \citep{bar2021detreg}           & CVPR'22      & 14.58$\pm$0.30             & 24.80$\pm$0.20             & 29.12$\pm$0.20            \\
MUM   \citep{kim2022mum}                 & CVPR'22      & \textbf{21.88}$\pm$0.12    & 28.52$\pm$0.09             & 31.87$\pm$0.30            \\ \midrule

Our Supervised Baseline                        & -              & 11.31$\pm$0.30             & 21.33$\pm$0.20             & 26.34$\pm$0.10     \\
Ours                                     & -              & 17.36$\pm$0.22             & \underline{29.84}$\pm$0.21 & \textbf{35.08}$\pm$0.34     \\

\toprule
\end{tabular}
\caption{\footnotesize{\textbf{Comparison with State-of-the-Art Methods on MS-COCO val2017 with PLD Setting.} {The mAP$_{50:95}$ standard COCO evaluation metrics on the COCO validation set are reported by models trained on 1, 5, 10\% Labelled Ratio under PLD settings. The results are the average of 5 experiments with different random seeds. Our supervised baseline refers to our model without the unlabelled branch, leaving a Deformable DETR setup with ResNet-50 backbone identically initialised to our full method for fair comparison. Note that our full method demonstrates competitive or superior performance in comparison, indicating that concepts introduced here for wildlife detection are still applicable to  general object detection.}}}
\label{tab:coco_pld}
\end{table*}

\begin{table}[ht]
\centering
\begin{tabular}{l|l|c}
\toprule
\textbf{Method}                          & \textbf{Venue} & \textbf{mAP} \\
\midrule \midrule
STAC \citep{sohn2020simple}              & Arxiv'20     & 39.21               \\
ISMT \citep{yang2021interactive}         & CVPR'21      & 39.64              \\
Unbiased-teacher \citep{liu2021unbiased} & ICLR'21      & 41.30               \\
Humble-Teacher \citep{tang2021humble}    & CVPR'21      & 42.37               \\
Instant-Teaching \citep{zhou2021instant} & CVPR'21      & 40.20               \\
Soft-Teacher \citep{xu2021end}           & ICCV'21      & 44.50                     \\
MUM \citep{kim2022mum}                   & CVPR'22      & 42.11                     \\
\midrule
Ours               & -     & \textbf{45.30}                     \\
\toprule
\end{tabular}
\caption[Comparison with State-of-the-Art Methods on MS-COCO val2017 with FLD Setting.]{\footnotesize{\textbf{Comparison with State-of-the-Art Methods on MS-COCO val2017 with FLD Setting.}} The mAP$_{50:95}$ standard COCO evaluation metrics on the COCO validation set are reported by models trained on all the labelled \texttt{train2017} set plus additional unlabelled \texttt{unlabeled2017}. Our method dominates SOTA methods.}
\label{tab:coco_fld}
\end{table}

\section{Experiments on the MS-COCO Dataset}
Our method primarily addresses the problem of handling sparsity of labelled data in animal biometrics whenever large unlabelled data is available. Yet, it is nevertheless both conceptually and practically applicable to mainstream object detection. The concept of slowly expanding detection capabilities of a model in a policy-controlled way to learn highly complex and variable object appearance is indeed not limited to animal detection.
In order to experimentally support any claim of wider applicability, we next evaluated our proposed method on the popular MS-COCO dataset under a low data regime (PLD) {and with extra unlabelled data (FLD)}.
For a fair comparison, we followed the evaluation approach used by STAC \citep{sohn2020simple} using their splits between labelled and unlabelled data {for PLD settings}. We trained our model with Labelled Ratios of 1\%, 5\%, and 10\% evaluated on the standard COCO \texttt{val2017} with the mAP$_{50:95}$ metrics.
{For the FLD option, we trained our model using the fully labelled COCO \texttt{train2017}, plus additional unlabelled COCO \texttt{unlabeled2017} following the same procedure described in} \citep{sohn2020simple,yang2021interactive,liu2021unbiased,tang2021humble,zhou2021instant,xu2021end,kim2022mum}.
As shown in Table~\ref{tab:coco_pld} and Table \ref{tab:coco_fld}, we achieve leading state-of-the-art results for a 10\% PLD Labelled Ratio {and the FLD setting}. At other ratios, our benchmarks remain competitive: for 5\% PLD our method trails only $0.90\%$ below the best result by SoftTeacher, and for 1\% PLD it scores $4.52\%$ below the SOTA MUM model. This demonstrates that the introduced concepts of dynamic control in curriculum learning are certainly applicable to a wider domain of general object detection.
We find that our curriculum learning method is less sensitive to hyper-parameters. In practice, the hyper-parameter configurations for COCO dataset\footnote{To handle the large size of MS-COCO, the training epochs are adjusted to range from 50 to 100 depending on the labelled ratio so that the total training iteration is fixed to 180k, while keeping the other hyper-parameters of the policy the same.} are inherited from the hyper-parameters that are fine-tuned on the PanAfrica dataset.
They can indeed outperform the state-of-the-art under certain configuration after searching. Further research will be required to stipulate in how far truly dataset-optimal hyper-parameterisation of dynamic training regimes such as the one presented is computationally feasible. For practical purposes, it is important to note that hyper-parameter transfer does not lead to learning collapse or vastly degraded performance as will be shown again in our experiments outlined in the next section.

\begin{table*}
\centering
\small
\begin{tabular}{llc|ll|ll}
\toprule
\multirow{2}{*}{No.} &\multirow{2}{*}{\textbf{Method}} & \multirow{2}{*}{\textbf{Venue}} & \multicolumn{2}{c|}{\textbf{VOC12}} & \multicolumn{2}{c}{\textbf{VOC12$+$COCO20cls}} \\
			&	 & & mAP$_{50}$                 & mAP  & mAP$_{50}$                 & mAP              \\
\midrule \midrule
1. & Supervised          & -          & $72.63$                              & $42.13$                              & $72.63$                             & $42.13$            \\
\midrule
2. & STAC    \citep{sohn2020simple}              & Arxiv'20 & $77.45$ \gain{\scriptsize{(+4.82)}}  & $44.64$ \gain{\scriptsize{(+2.51)}}  & $79.08$ \gain{\scriptsize{(+6.45)}} & $46.01$ \gain{\scriptsize{(+3.88)}}          \\
3. &ISMT    \citep{yang2021interactive}         & CVPR'21  & $77.23$ \gain{\scriptsize{(+4.60)}}  & $46.23$ \gain{\scriptsize{(+4.10)}}  & $77.75$ \gain{\scriptsize{(+5.12)}} & $49.59$ \gain{\scriptsize{(+7.46)}}         \\
4. &Instant-Teaching  \citep{zhou2021instant}   & CVPR'21  & $79.20$ \gain{\scriptsize{(+6.57)}}  & $50.00$ \gain{\scriptsize{(+7.87)}}  & $79.00$ \gain{\scriptsize{(+6.37)}} & $50.80$ \gain{\scriptsize{(+8.67)}}           \\
5. &Humble-Teacher  \citep{tang2021humble}      & CVPR'21  & $80.94$ \gain{\scriptsize{(+8.31)}}  & $53.04$ \gain{\scriptsize{(+10.91)}} & $81.29$ \gain{\scriptsize{(+8.66)}} & $54.41$ \gain{\scriptsize{(+12.28)}}           \\
6. &Unbiased-Teacher \citep{liu2021unbiased}    & CVPR'21  & $77.37$ \gain{\scriptsize{(+4.74)}}  & $48.69$ \gain{\scriptsize{(+6.56)}}  & $78.82$ \gain{\scriptsize{(+8.19)}} & $50.34$ \gain{\scriptsize{(+8.21)}}          \\
7. &Unbiased-Teacher-v2 \citep{liu2022unbiased} & CVPR'22  & $81.29$ \gain{\scriptsize{(+8.66)}}  & $56.87$ \gain{\scriptsize{(+14.74)}} & $82.04$ \gain{\scriptsize{(+9.41)}} & $58.08$ \gain{\scriptsize{(+15.95)}}           \\
8. &MUM         \citep{kim2022mum}              & CVPR'22  & $78.94$ \gain{\scriptsize{(+6.31)}}  & $50.22$ \gain{\scriptsize{(+8.09)}}  & $80.45$ \gain{\scriptsize{(+7.82)}} & $52.31$ \gain{\scriptsize{(+10.18)}}     \\
9. &LabelMatch     \citep{chen2022label}        & CVPR'22  & $\mathbf{85.48}$ \gain{\scriptsize{(+12.85)}} & $55.11$ \gain{\scriptsize{(+12.98)}} & -                                   & -          \\
10. &DSL         \citep{chen2022dense}           & CVPR'22  & $80.70$ \gain{\scriptsize{(+8.07)}}  & $\underline{56.80}$ \gain{\scriptsize{(+14.67)}} & $\underline{82.10}$ \gain{\scriptsize{(+9.47)}} & $\mathbf{59.80}$ \gain{\scriptsize{(+17.67)}}    \\
11. &ACRST        \citep{zhang2022semi}          & AAAI'22  & $81.11$ \gain{\scriptsize{(+8.48)}}  & $54.30$ \gain{\scriptsize{(+12.17)}} & -                                   & -   \\
12. &Dense-Teacher    \citep{zhou2022dense}      & ECCV'22  & $79.89$ \gain{\scriptsize{(+7.26)}}  & $55.87$ \gain{\scriptsize{(+13.74)}} & $81.23$ \gain{\scriptsize{(+8.60)}} & $57.52$ \gain{\scriptsize{(+15.39)}}           \\
\midrule
13. &Ours$^{\dagger}$ &  & $81.89$ \gain{\scriptsize{(+9.26)}}             & $57.02$ \gain{\scriptsize{(+14.89)}}  & $81.82$ \gain{\scriptsize{(+9.19)}} & $58.28$ \gain{\scriptsize{(+16.15)}}           \\
14. &Ours & &  $\underline{82.09}$ \gain{\scriptsize{(+9.46)}}            &  $\mathbf{57.65}$ \gain{\scriptsize{(+15.52)}} & $\mathbf{82.34}$ \gain{\scriptsize{(+9.71)}}            &  $\underline{58.85}$ \gain{\scriptsize{(+16.72)}}\\
\bottomrule
\end{tabular}
\caption{\footnotesize{\textbf{Comparison on PASCAL VOC Dataset.}
		In experiment, \texttt{VOC2007-trainval} is used as the labelled set and \texttt{VOC2012-trainval} used as the unlabelled set for all the models. The results are reported based on the evaluation on \texttt{VOC2007-test}. ${\dagger}$ represents the hyperparameters that are directly inherited from COCO without further finetuning efforts.
}}
\label{tab:voc_restult}
\end{table*}

\begin{table*}
\centering
\small
\begin{tabular}{lc|lll}
\toprule
\textbf{Method}       & \textbf{Labelled Ratio} & \textbf{mAP }                                        & mAP$_{50}$                                           & mAP$_{75}$                                           \\
\midrule \midrule
Supervised baseline   & $5\%$                & 26.15$\pm$1.47                                       & 65.24$\pm$2.55                                       & 14.96$\pm$1.83                                       \\
Ours                  & $5\%$                & \textbf{32.81}$\pm$1.40 \gain{\scriptsize{(+6.66)}} & \textbf{73.19}$\pm$2.93 \gain{\scriptsize{(+7.95)}}  & \textbf{22.12}$\pm$0.89 \gain{\scriptsize{(+7.16)}}  \\
\midrule
Supervised   baseline & $10\%$               & 35.40$\pm$1.15                                       & 75.82$\pm$1.31                                       & 27.16$\pm$1.91                                       \\
Ours                  & $10\%$               & \textbf{40.47}$\pm$0.25 \gain{\scriptsize{(+5.07)}}  & \textbf{79.15}$\pm$0.60  \gain{\scriptsize{(+3.33)}} & \textbf{35.83}$\pm$1.05  \gain{\scriptsize{(+8.67)}} \\
\midrule
Supervised  baseline  & $20\%$               & 42.24$\pm$1.33                                       & 82.34$\pm$1.65                                       & 37.99$\pm2.12$                                       \\
Ours                  & $20\%$               & \textbf{45.17}$\pm$1.02 \gain{\scriptsize{(+2.93)}}  & \textbf{83.79}$\pm$0.89 \gain{\scriptsize{(+1.45)}}  & \textbf{44.09}$\pm$0.85 \gain{\scriptsize{(+6.10)}}  \\
\bottomrule
\end{tabular}
\caption{\footnotesize{\textbf{Experimental Results on Bees Dataset.} Mean and standard deviation on test set portion evaluated over 5 data folds for 5\%, 10\%, 20\% labelled ratio are reported. Supervised baseline refers to the same model trained on the labelled data only. Note that we adopt the policy hyper-parameters optimised for the PanAfrica dataset on these experiments.}}
\label{tab:bee_restult}
\end{table*}

\section{Experiments on the PASCAL VOC Dataset}
In order to understand applicability to mainstream object detection further, we utilise another popular object detection benchmark to evaluate our model.
We follow the standard FLD evaluation process on the PASCAL VOC dataset (\cite{everingham2010pascal}), as in \citep{sohn2020simple,liu2021unbiased,zhou2021instant}, with the performance of our model reported on \texttt{VOC07-test}, trained using \texttt{VOC07-trainval} as the labelled training set, and \texttt{VOC12-trainval} or \texttt{VOC12-trainval} + \texttt{COCO20cls}\footnote{This set is tailored from MS-COCO dataset, which keeps the same 20 categories as PASCAL VOC as the unlabelled training set.}.

As shown in Table \ref{tab:voc_restult}, we explore two different policy-parameter settings in the experiments, i) without policy-parameter searching\footnote{We heuristically adopt the policy-parameter fine-tuned for MS-COCO.} (with $\dagger$ notation in the Table), and ii) with pseudo-label analysis and policy-parameter searching. For VOC12,  Row 13 shows the leading 57.02\% mAP and the second best 81.89 \% mAP$_{50}$, and for VOC12 + COCO20cls, Row 13 offers the second best 58.28\% mAP and competitive 81.82\% mAP$_{50}$ among state-of-the-art methods. It also achieves a 14.89\% and 16.15\% gain in mAP over the supervised baselines, respectively, by simply adopting the configuration from the MS-COCO experiments. This further supports the argument that policy and parameter transfer does not lead to learning collapse or vast performance degradation. 

When we systematically analyse the pseudo-labels and perform policy-hyper-parameter finetuning\footnote{For reproducibility of this experiment, the exact parameters used were: $\pi$ Linear Increase with warm-up and cool-down phases;  linear increase $\alpha~ 0.1 \rightarrow 1$; arctan increase $\varsigma~ 0.2 \rightarrow 0.5$.}, the performance of our method can be  boosted, achieving state-of-the-art $57.65\%$ mAP for VOC12  and $82.34\%$ mAP$_{50}$ for VOC12 + COCO20cls (row 14 in Table \ref{tab:voc_restult}).

Our experimental results on PASCAL VOC suggest i) the proposed method can have applications beyond animal detection, ii) it does not need heuristic tuning for hyper-parameters, since merely adopting the COCO ones to PASCAL VOC can {lead to virtuous training cycles and} achieve  {competitive} results.

\begin{table*}[ht]
	\centering \vspace{15pt}
	\begin{tabular}{lc|ccc|ccc}
\toprule
\multirow{2}{*}{\textbf{Labelled}}& \multirow{2}{*}{\textbf{fold}}  &\multicolumn{3}{c|}{\textbf{Supervised Baseline}}  &\multicolumn{3}{c}{ \cellcolor[HTML]{EFEFEF}  \textbf{Ours}}  \\
				  &		     & \textbf{mAP} & \textbf{mAP}$_{50}$ & \textbf{mAP}$_{75}$ & \cellcolor[HTML]{EFEFEF}\textbf{mAP} & \cellcolor[HTML]{EFEFEF}\textbf{mAP}$_{50}$ & \cellcolor[HTML]{EFEFEF}\textbf{mAP}$_{75}$ \\
						     \midrule \midrule
\multirow{4}{*}{5\%}  & 0             & 52.13 & 74.94 & 56.45 & \cellcolor[HTML]{EFEFEF} 55.40 & \cellcolor[HTML]{EFEFEF} 78.48 & \cellcolor[HTML]{EFEFEF} 59.20 \\
                      & 1             & 53.31 & 75.75 & 57.34 & \cellcolor[HTML]{EFEFEF} 56.11 & \cellcolor[HTML]{EFEFEF} 78.16 & \cellcolor[HTML]{EFEFEF} 60.18 \\
                      & 2             & 52.05 & 74.98 & 56.12 & \cellcolor[HTML]{EFEFEF} 55.38 & \cellcolor[HTML]{EFEFEF} 78.17 & \cellcolor[HTML]{EFEFEF} 59.61 \\
                      & \textit{avg.} & 52.50 & 75.22 & 56.64 & \cellcolor[HTML]{EFEFEF} 55.63 & \cellcolor[HTML]{EFEFEF} 78.27 & \cellcolor[HTML]{EFEFEF} 59.66 \\
 \midrule
\multirow{4}{*}{10\%} & 0             & 56.24 & 78.00 & 60.83 & \cellcolor[HTML]{EFEFEF} 59.56 & \cellcolor[HTML]{EFEFEF} 80.92 & \cellcolor[HTML]{EFEFEF} 63.86 \\
                      & 1             & 55.96 & 78.46 & 60.95 & \cellcolor[HTML]{EFEFEF} 59.14 & \cellcolor[HTML]{EFEFEF} 80.75 & \cellcolor[HTML]{EFEFEF} 63.47 \\
                      & 2             & 55.89 & 78.87 & 60.81 & \cellcolor[HTML]{EFEFEF} 58.95 & \cellcolor[HTML]{EFEFEF} 80.61 & \cellcolor[HTML]{EFEFEF} 63.60 \\
                      & \textit{avg.} & 56.03 & 78.44 & 60.86 & \cellcolor[HTML]{EFEFEF} 59.22 & \cellcolor[HTML]{EFEFEF} 80.76 & \cellcolor[HTML]{EFEFEF} 64.33 \\
\midrule
\multirow{4}{*}{20\%} & 0             & 58.98 & 81.19 & 64.26 & \cellcolor[HTML]{EFEFEF} 60.97 & \cellcolor[HTML]{EFEFEF} 82.53 & \cellcolor[HTML]{EFEFEF} 66.04 \\
                      & 1             & 59.21 & 80.96 & 64.17 & \cellcolor[HTML]{EFEFEF} 61.39 & \cellcolor[HTML]{EFEFEF} 82.78 & \cellcolor[HTML]{EFEFEF} 66.32 \\
                      & 2             & 59.74 & 81.51 & 64.81 & \cellcolor[HTML]{EFEFEF} 62.12 & \cellcolor[HTML]{EFEFEF} 82.90 & \cellcolor[HTML]{EFEFEF} 67.26 \\
                      & \textit{avg.} & 59.31 & 81.22 & 64.41 & \cellcolor[HTML]{EFEFEF} 61.49 & \cellcolor[HTML]{EFEFEF} 82.74 & \cellcolor[HTML]{EFEFEF} 66.54 \\

\midrule
	\end{tabular}
	\caption{
		\footnotesize{\textbf{Experimental Results on Snapshot Serengeti Dataset.}
			Three folds and their average results on test set evaluated are reported for 5\%, 10\%, 20\% labelled ratio. Supervised baseline refers
			to our model without any aspect of the unlabelled branch. Note that we adopt the policy hyper-parameters optimised for the PanAfrica dataset on these experiments.\vspace{15pt}
		}
	}
\label{tab:snap_restult}
\end{table*}

\section{Experiments on the Bees Dataset}
The Bees dataset\footnote{Available at \url{https://lila.science/datasets/boxes-on-bees-and-pollen}} contains approximately 5K images of bees captured in hives. The bees and pollen that appear in each image are annotated with Bounding boxes and most of the data includes crowded scenes where bees are densely located.

In this experiment, we consider only PLD settings as we do not have extra bee data.
We randomly spilt 80\% of the whole data as a training set and use the rest as a testing set.
For the training set, we construct three different PLD labelled ratios sampled with five different random seeds, where the labels are randomly masked so that the proportion of labelled data is 5\%, 10\%, and 20\%, respectively.
We evaluate the supervised baseline (using the same baseline approach as for MS-COCO) and our proposed model over five data folds for 5\%, 10\%, and 20\% labelled ratios and report the mean and standard deviation of mAP.
In Table \ref{tab:bee_restult}, we demonstrate a substantial performance boost by applying our policy-guided semi-supervised learning, especially under lower data regimes, with $6.66\%$ gain over the baseline in the 5\% PLD setting.

\section{Experiments on the Snapshot Serengeti Dataset}
We finally conducted experiments on sparsely labelled versions of the Snapshot Serengeti dataset\footnote{Available at \url{https://lila.science/datasets/snapshot-serengeti}} (\cite{swanson2015snapshot}) in which overall around 78K images (out of 7.1M)
are labelled with instance-level bounding boxes that allow us to test our proposed method.  We conducted our experiments under PLD settings where the model was trained with 5\% and 10\%  of the labelled data out of 78K labelled images.

As shown in Table \ref{tab:snap_restult}, substantial boosts of mAP, mAP$_{50}$, mAP$_{75}$ can be observed under limited label regimes when comparing the supervised baseline (composed as before for the MS-COCO and Bees datasets) to our full system. Moreover, our method can reach similar or better performance than the supervised baseline while using only half of the labelled data.

To provide some further context to the wider literature on this dataset, we note that using only $20\%$ of the labels our method's performance at $mAP_{50}$ of 82.7 comes close to published results for fully supervised training with $100\%$ of the Snapshot Serengeti labels using Mask-RCNN~\citep{ibraheam2021performance} at $mAP_{50}$ of 85.7 and significantly outperforms full label training with Faster-RCNN~\cite{ibraheam2021performance} at $mAP_{50}$ of 73.2 or Context-RCNN~\citep{beery2020context} at $mAP_{50}$ of 55.9.

Finally, we note that the multi-dataset learning approach of the MegaDetectorV5a*~\citep{megadetector} leading to $mAP_{50}$ of 90.65 on this dataset prevents fair apple-to-apple comparison with our method. However, the multi-dataset training regime is clearly highly effective in utilising label information across animal species boundaries. Future work into benchmarking our presented work in such multi-dataset training settings seems a promising avenue to improve results for species-specific and species-agnostic detectors further.

\section{Conclusion}
In this paper, we introduced an end-to-end dynamic curriculum learning framework for semi-supervised detection in sparely labelled datasets unlocking information in the unlabelled data portions. We demonstrated that bipolarity in the behaviour of cyclical student-teacher training regimes can lead to either effective virtuous or collapsing vicious training loops. We discussed the importance of expanding model coverage of new data slowly and in a controlled way to keep expanding detector and label quality without collapse. To achieve this, we proposed five policies to guide the dynamics of training and promote steady, simultaneous improvements to the student detector, the teacher detector, and the quality of the pseudo-labels. We showed that the described approach is effective in significantly advancing the state-of-the-art in great ape detection performance when evaluated under various settings on the large Extended PanAfrican Dataset. Our method is also shown to be beneficial  to sparse labelling versions of other datasets without specialising hyper-parameterisations or policies. We have demonstrated this for the Bees and Snapshot Serengeti datasets in the animal domain. Finally, we showed that evaluation on general object detection tasks in MS-COCO and PASCAL-VOC achieves competitive or superior performance over existing state-of-the-art methods.

We conclude that the work holds the promise for dynamic curriculum learning controlled by training policies to be applied effectively to  sparsely labelled wildlife data and thereby help unlock the full wealth of information so far widely sealed in steadily growing unlabelled camera trap archives.
\newpage

\section*{Acknowledgements}
\label{sec:ack}
We would like to thank the entire team of the Pan African Programme: ‘The Cultured Chimpanzee’~\cite{mpi} and its collaborators for allowing the use of their data for this project. Please contact the copyright holder Pan African Programme at \url{http://panafrican.eva.mpg.de} to obtain the source videos from the dataset. Particularly, we thank: H Kuehl, C Boesch, M Arandjelovic, and P Dieguez. We would also like to thank: K Zuberbuehler, K Corogenes, E Normand, V Vergnes, A Meier, J Lapuente, D Dowd, S Jones, V Leinert, EWessling, H Eshuis, K Langergraber, S Angedakin, S Marrocoli, K Dierks, T C Hicks, J Hart, K Lee, and M Murai. Thanks also to the team at \url{https://www.chimpandsee.org}. The work that allowed for the collection of the dataset was funded by the Max Planck Society, Max Planck Society Innovation Fund, and Heinz L. Krekeler. In this respect we would also like to thank: Foundation Ministre de la Recherche Scientifique, and Ministre des Eaux et Forłts in Cote d’Ivoire; Institut Congolais pour la Conservation de la Nature and Ministre de la Recherch Scientifique in DR Congo; Forestry Development Authority in Liberia; Direction des Eaux, Forłts Chasses et de la Conservation des Sols, Senegal; and Uganda National Council for Science and Technology, Uganda Wildlife Authority, National Forestry Authority in Uganda.

\bibliography{main}
\end{document}